\documentclass[11pt]{article}

\usepackage[final]{acl}
\usepackage{booktabs}
\usepackage{graphicx}
\usepackage{enumitem}
\usepackage{times}
\usepackage{latexsym}
\usepackage{amsmath}
\usepackage{amsfonts}
\usepackage{stfloats}
\usepackage{tabularx}
\usepackage{algorithm}
\usepackage{algpseudocode}  
\usepackage{tcolorbox}
\tcbuselibrary{skins, breakable} % <-- 加载 skins 和 breakable 两个库
\usepackage{booktabs}   % For professional-looking rules (\toprule, \midrule, \bottomrule)
\usepackage{multirow}   % For spanning multiple rows
\usepackage{graphicx}   % For \resizebox
\usepackage[T1]{fontenc}
\usepackage[utf8]{inputenc}

\usepackage{microtype}

\usepackage{inconsolata}

\usepackage{graphicx}

\title{EvolMathEval: Towards Evolvable Benchmarks for Mathematical Reasoning via Evolutionary Testing}

\author{
  Shengbo Wang\textsuperscript{1},
  Mingwei Liu\textsuperscript{1}\thanks{\ \ Corresponding Author: liumw26@mail.sysu.edu.cn},
  Zike Li\textsuperscript{1},
  Anji Li\textsuperscript{1}, 
  Yanlin Wang\textsuperscript{1},
  Xin Peng\textsuperscript{2},
  Zibin Zheng\textsuperscript{1} \\
  \textsuperscript{1}Sun Yat-sen University \quad
  \textsuperscript{2}Fudan University \\
  \texttt{puppytagge@gmail.com, lizk8@mail2.sysu.edu.cn, lianj8@mail2.sysu.edu.cn,}\\
  \texttt{wangylin36@mail.sysu.edu.cn, zhzibin@mail.sysu.edu.cn, pengxin@fudan.edu.cn}
}

\begin{document}
\maketitle
\begin{abstract}
The rapid advancement of Large Language Models (LLMs) poses a significant challenge to existing mathematical reasoning benchmarks. 
% However, these benchmarks commonly suffer from issues such as score saturation, temporal decay, and data contamination. 
However, these benchmarks tend to become easier over time as LLMs can learn from the published benchmarks.
This limitation hinder the precise evaluation of the true capabilities of SOTA models. 
To address this challenge, this paper introduces EvolMathEval, an automated mathematical benchmark generation and evolution framework based on evolutionary testing. 
% By dynamically generating unique evaluation instances, the framework ensures the benchmark remains perpetually challenging for future models.
% The core mechanisms of EvolMathEval include: seed problem generation based on reverse engineering with algebraic guarantees; multi-dimensional genetic operators designed to inject diverse cognitive challenges; and a composite fitness function that can rapidly and accurately assess problem difficulty. 
% the proposed composite fitness function can efficiently and precisely quantify the difficulty of mathematical problems. 
Experimental results demonstrate that EvolMathEval can not only generate a large volume of high-difficulty problems through continuous self-iteration, but it can also significantly enhance the complexity of public datasets like GSM8K through evolution, reducing model accuracy by an average of 48\%.
Deeper investigation reveals that when solving these evolved problems, LLMs tend to bypass complex multi-step logical reasoning by relying on simplistic and fuzzy conditions, consequently leading to incorrect solutions. We define this phenomenon as the ``Pseudo Aha Moment", which we find accounts for 77\% to 100\% of errors on targeted problems.
% This finding uncovers a cognitive shortcut-taking behavior in the deep reasoning processes of current LLMs, which we find accounts for 77\% to 100\% of errors on targeted problems.
% This finding reveals a tendency for current LLMs to take shortcuts during deep reasoning, which accounts for 77\% to 100\% of errors on targeted problems.
Code and resources are available at: https://anonymous.4open.science/r/EvolMathEval
\end{abstract}

\section{Introduction}

Mathematical reasoning serves as a cornerstone for evaluating the capabilities of LLMs \cite{ahn-etal-2024-large}, yet established static benchmarks such as GSM8K \cite{cobbe2021trainingverifierssolvemath} and MATH \cite{hendrycksmath2021} are facing a validity crisis. Over time, these datasets may become easier for LLMs or be incorporated into the training corpora for the latest LLMs, which can then achieve near-perfect results \cite{balloccu-etal-2024-leak, dong-etal-2024-generalization, Jiang2024InvestigatingDC, sainz-etal-2023-nlp}. 
% More critically, they are compromised by \textbf{data contamination}, which allows models to attain high scores by memorizing answers leaked into training data rather than engaging in genuine reasoning \cite{Balloccu2024LeakCR, Dong2024GeneralizationOM, Jiang2024InvestigatingDC}. 
As revealed by recent studies on models like Qwen2.5 \cite{wu2025reasoningmemorizationunreliableresults}, this reliance on memorization means that high scores no longer reliably indicate true problem-solving ability. 
The fundamental problem is that these benchmarks are static.

% The static nature of these benchmarks is the fundamental vulnerability that allows this cycle of contamination and saturation to persist.
% The fundamental problem is that these benchmarks are static, which allows the cycle of contamination and saturation to continue.

To overcome the limitations of static benchmarks, the research focus has shifted toward dynamic generation. 
However, existing methods encounter their own bottlenecks. 
Approaches that leverage LLMs to directly generate problems \cite{Liu2024Agents4PLCAC, Zhang2025DeployabilityCentricIG, Lee2024VISTAVI} often struggle with precise control over quality and difficulty. Meanwhile, programmatic or template-based solutions \cite{Xie2024AdversarialMW, fan-etal-2024-nphardeval, kurtic-etal-2024-mathador} are prone to homogenization in their logical structures, lacking the diversity required to challenge authentic reasoning abilities. 
A critical flaw in most dynamic frameworks is their lack of an efficient, and automated evaluation and feedback loop. 
Consequently, they cannot iteratively optimize for both difficulty and quality.\cite{Kang2024TemplateDrivenLF, Wu2022AutomaticMW, liu-etal-2024-mathbench}. 
Therefore, the core challenge in this domain is how to simultaneously achieve diversity, and controllable difficulty within a single framework.

% 看到这里
To address this challenge, we propose EvolMathEval, an automated framework that utilizes evolutionary testing \cite{Chi2010IntroductionAS} to dynamically generate and evolve mathematical reasoning benchmarks. 
Its process begins by generating initial ``seed" problems from scratch via reverse-engineering. 
Subsequently, it applies a series of genetic operators to evolve either these new problems or those from existing datasets.
A fitness function is then used to evaluate the difficulty of the problems. 
Ultimately, we use secondary evolution to increase the difficulty of simpler problems.
This adaptive, closed-loop mechanism ensures that EvolMathEval can generate novel and challenging problems that models have not previously encountered. 
We name the benchmark generated by this framework EvolMath.

Experiments demonstrate that EvolMath effectively challenges SOTA LLMs.
Even the top-performing model (DeepSeek-R1), achieved only 79\% accuracy on EvolMath. 
When EvolMathEval is applied to public datasets, it reduces model accuracy by 48\% on average and up to 95\% in some cases. 
Furthermore, problems generated via secondary evolution can even cause the accuracy of some models to drop to zero.
This performance degradation uncovers capability differences that were masked by the previous benchmarks.

Critically, our failure analysis on these evolved problems uncovers a key weakness we term the \textbf{``Pseudo Aha Moment"}: 
Models tend to favor simplistic, logically flawed ``shortcuts" over rigorous, multi-step reasoning. 
% Our analysis reveals that this phenomenon is highly prevalent among incorrect solutions, with its incidence rate ranging from a minimum of 77\% to as high as 100\% for some models.
Our analysis reveals this phenomenon is a major cause of errors in the target problems. 
% In fact, for some models, it was responsible for anywhere from 77\% to all (100\%) of their incorrect answers.
In fact, its responsibility for incorrect answers ranged from 77\% to 100\% across the different LLMs tested.

Our contributions are as follows:
\begin{itemize}[leftmargin=*,topsep=0pt,itemsep=-2pt]
% \item We propose EvolMathEval, which provides a solution to the widespread problems of score saturation and data contamination in static benchmarks by dynamically and perpetually generating novel problems.

\item We propose a framework named EvolMathEval, which utilizes evolutionary testing to continuously and dynamically generate novel and challenging problems. This framework can create problems either from scratch or by evolving existing benchmarks.

% \item We design and implement a set of genetic operators and a data-driven composite fitness function. The former systematically increases problem difficulty, while the latter enables the precise quantification and automated selection of problems, collectively driving the iterative evolution of the entire benchmark.

\item We introduce EvolMath, a new benchmark generated using our EvolMathEval framework. It can more clearly measure the true mathematical reasoning capabilities of different LLMs.

\item We are the first to discover and quantify the ``Pseudo Aha Moment" across mainstream LLMs. 
This finding reveals a deep weakness in the mathematical capabilities of current LLMs.
\end{itemize}

\section{Related Work}

\textbf{Evaluating Mathematical Reasoning in LLMs:} 
The evaluation of mathematical reasoning in LLMs follows two approaches: 
Assessing the final answer's accuracy or analyzing the reasoning process. 
Standard benchmarks like GSM8K and MATH \cite{patel-etal-2021-nlp} exemplify the first approach, while methods like Chain of Thought supervision \cite{NEURIPS2022_9d560961, zhang2023automatic, Mitra2023CompositionalCP} are used for the second.
% The evaluation of mathematical reasoning in LLMs relies on either assessing answer accuracy on standard benchmarks like GSM8K and MATH \cite{patel-etal-2021-nlp} or analyzing the reasoning process itself, for instance, through Chain of Thought supervision \cite{NEURIPS2022_9d560961, zhang2023automatic, Mitra2023CompositionalCP}. 
However, the validity of these methods usually relies on static benchmarks, which lose their effectiveness over time.

\textbf{Evolutionary Testing:} To address this, we introduce Evolutionary Testing \cite{5954405}, an automated method from software engineering. 
In EvolMathEval, we treat problems as "test cases" and their difficulty as ``fitness", allowing us to continuously evolve novel and challenging problems via genetic operators. 
It ensures the reasoning process is not a reproduction of memorized solutions.

\section{Methodology}
\label{sec:evolreasonaval_framework} % Added a label for cross-referencing

EvolMathEval is an automated framework for mathematical reasoning benchmark generation and evolution, grounded in the theory of evolutionary testing. 
The entire framework is composed of four core modules: Seed Problem Initialization, Genetic Operators, Crossover Operator, and the Fitness Function.
% Figure~\ref{fig:shortcut_variable} shows the EvolMathEval workflow.
Figure~\ref{fig:shortcut_variable} shows the EvolMathEval workflow, with full details provided in Appendix~\ref{DetailsofEvolMathEval}.

% \begin{figure}[h]
%     \centering
%     % \includegraphics[width=\textwidth]{figures/EvolMathEval_Framework_Flowchart.png}
%     \includegraphics[width=\columnwidth]{latex/figures/new_framework.png}
%         \caption{An overview of the EvolMathEval framework.}
%     \label{fig:shortcut_variable}
% \end{figure}

\begin{figure*}[h]
    \centering
    \includegraphics[width=\textwidth, trim=0cm 0cm 0cm 2.1cm, clip]{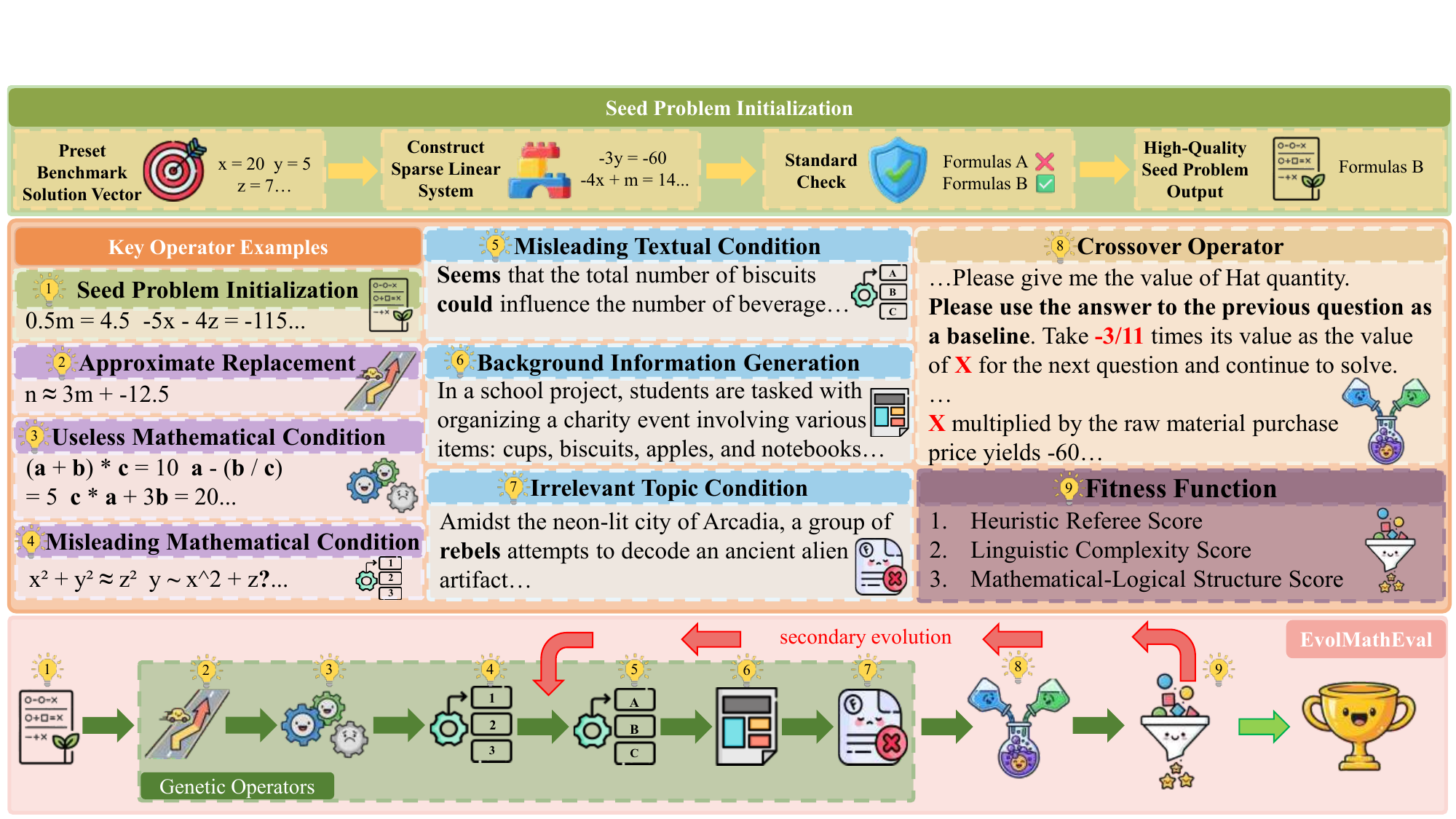}
        \caption{An overview of the EvolMathEval framework, illustrating the Seed Problem Initialization stage (top), examples of Key Operators (middle), and the complete evolutionary framework (bottom).}
    \label{fig:shortcut_variable}
\end{figure*}

\subsection{Seed Problem Initialization}
\label{Seed_Problem_Initialization}

% \begin{figure}[h]
%     \centering
%     \includegraphics[scale=0.7]{latex/figures/Seed_Problem.png}
%         \caption{Workflow of the Seed Problem Initialization.}
%     \label{fig:Seed_Problem}
% \end{figure}

EvolMathEval begins with Seed Problem Initialization, which procedurally generates the initial population of problems for evolution. 
To ensure each problem is logically rigorous and has a guaranteed, unique solution, the core approach is to work backward from the answer. 
We begin by pre-setting the correct solution, then construct a series of simple mathematical conditions around it. 
Each condition connects only a few variables, encouraging step-by-step reasoning. 
Finally, every problem undergoes rigorous algebraic checks to validate its quality and consistency. 
% Figure~\ref{fig:Seed_Problem} shows the workflow.
See the workflow at the top of Figure~\ref{fig:shortcut_variable}.
The detailed methodology for this generation process, including the specific mathematical guarantees, is provided in Appendix~\ref{app:Seed_Problem}.

\subsection{Genetic Operators}
After the initial population is generated, genetic operators are applied to enhance the diversity of the population. 
To ensure both the accuracy of the mathematical core and the richness of the natural language expression, EvolMathEval's workflow strictly decouples algebraic manipulation from language generation. 

The process involves three stages:
First, formulaic mutation operators algebraically modify the core formulas to alter the problem's logical structure.
Second, these modified formulas are converted into a natural language draft via templates and subsequently polished by an LLM for fluency.
Finally, linguistic mutation operators inject narrative complexity and distractors into the polished text to challenge a model's comprehension.
% The details of formulaic mutation and linguistic mutation can be seen in Table~\ref{tab:mutation_operators_resizebox}.
Table~\ref{tab:mutation_operators_resizebox} details the mechanism of each operator, with corresponding visual examples provided in Appendix~\ref{app:full_example}.

\begin{table*}[tp]
\centering
\resizebox{\linewidth}{!}{%
\begin{tabular}{@{}p{1.5cm} p{1.5cm} p{14cm}@{}}
\toprule
\textbf{Category} & \textbf{Operator} & \textbf{Mechanism} \\
\midrule

% Formulaic Operators Section
\multirow{3}{*}{\textbf{Formulaic}} & \textbf{AR} & $S' = S \cup \{x_{\text{target}} \approx f(x_{\text{shortcut}})\}$: Adds a pseudo-condition linking a "shortcut variable" to the target. \\
\cmidrule(lr){2-3}
& \textbf{UMC} & $S' = S \cup S_{\text{useless}}$, s.t. $\text{vars}(S) \cap \text{vars}(S_{\text{useless}}) = \emptyset$: Adds a set of valid but logically decoupled equations involving a different set of variables. \\
\cmidrule(lr){2-3}
& \textbf{MMC} & $S' = S \cup \{g(x_i, x_j) \leadsto c\}$, where $x_i, x_j \in \text{vars}(S)$: Injects ambiguous, algebraically unsolvable conditions that involve variables from the main problem. \\
\midrule

% Linguistic Operators Section
\multirow{3}{*}{\textbf{Linguistic}} & \textbf{MTC} & $P' = P \cup \{T_{\text{false\_causality}}\}$: Adds descriptions with ambiguous references or false causal relationships. \\
\cmidrule(lr){2-3}
& \textbf{BI} & $P' = T_{\text{context}}(P_{\text{core}})$: Wraps the problem in a real-world scenario, adding narrative context. \\
\cmidrule(lr){2-3}
& \textbf{ITC} & $P' = P \cup T_{\text{irrelevant}}$, s.t. $\text{topic}(P) \cap \text{topic}(T_{\text{irr}}) = \emptyset$: Injects narrative fragments completely unrelated to the problem's central theme. \\
\bottomrule
\end{tabular}%
}
\caption{An overview of the genetic mutation operators. The abbreviations for the operators are as follows: 
\textbf{AR} (Approximate Replacement), 
\textbf{UMC} (Useless Math Condition), 
\textbf{MMC} (Misleading Math Condition), 
\textbf{MTC} (Misleading Textual Condition), 
\textbf{BI} (Background Information), and 
\textbf{ITC} (Irrelevant Topic Condition).}
\label{tab:mutation_operators_resizebox}
\end{table*}

\subsection{Crossover Operator}
\label{Crossover_Operator}

\begin{figure}[h]
    \centering
    \includegraphics[width=\columnwidth]{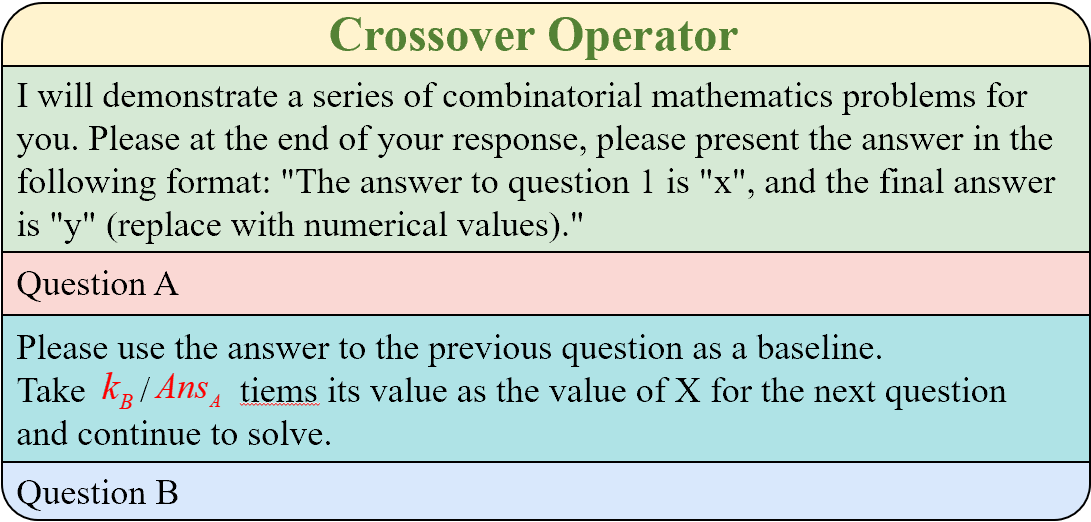}
        \caption{Framework of the Crossover Operator. The red variables will be replaced with the actual numerical value from each question.}
    \label{fig:Crossover_Operator}
\end{figure}

This operation aims to increase the length of the reasoning chain. It merges two independent ``parent" problems into sequential ``child" problems. 
% Its mechanism involves calculating a precise proportional relationship between key numerical values of the two problems and generating explicit natural language instructions. 
% This forces the model to complete a full, long-chain reasoning process of ``solve \(\rightarrow\) calculate \(\rightarrow\) re-solve". Figure~\ref{fig:Crossover_Operator} shows the framework.

The operator begins by retrieving the answer of the first parent problem ($Ans_A$) and identifying the initial numerical value ($k_B$) in the second parent problem's text. 
The value $k_B$ is then substituted with a placeholder, `X', in the second problem. 
Finally, a ratio is computed ($R = k_B / Ans_A$), and an instruction is generated that prompts the model to find the value of `X' by multiplying the first problem's answer by this ratio. 

This forces the model to complete a full, long-chain reasoning process of ``solve \(\rightarrow\) calculate \(\rightarrow\) re-solve". Figure~\ref{fig:Crossover_Operator} shows the framework.

\subsection{Fitness Function}
% The fitness function is analogous to the ``environment" in natural selection, responsible for the quantitative evaluation of each problem in the dataset to decide whether it requires a second round of evolution. 
% The function is a multi-dimensional, comprehensive evaluation model that extracts features from three levels:

The fitness function is the core selection mechanism in our evolutionary framework, designed to quantitatively assess the difficulty of each problem. 
It operates as a composite model, aggregating features from three dimensions: 
linguistic complexity, mathematical-logical complexity, and a heuristic difficulty score from a referee LLM. 
To handle the heterogeneous scales of these features, all feature values are normalized to a common range.
Problems with fitness scores below a predefined threshold are deemed "unfit" and are consequently selected for a subsequent round of evolution.
Table~\ref{tab:fitness_features} shows the details of the fitness features.

After extracting all features, we use a weight vector to calculate the comprehensive fitness score for each problem by the features. 
Notably, this weight vector is derived from a correlation and significance analysis of each metric.
This analysis is performed with respect to the model's evaluation results (a detailed analysis is presented in Section~\ref{sec:validation_fitness}).

% Finally, a dual-filter mechanism selects individuals, rejecting those with a low comprehensive fitness score or a score in the bottom 1st percentile on any single metric.

Finally, a dual-filter mechanism is employed for selection, rejecting individuals if their comprehensive fitness score falls below a threshold or if any single metric score is in the bottom 1st percentile.

% Finally, a dual-filter mechanism is employed for selection:
% \begin{itemize}
%     \item \textbf{Comprehensive Score Filter:} Rejects individuals whose comprehensive fitness score is below a preset threshold.
%     \item \textbf{Single-Metric Deficiency Filter:} Rejects any individual whose score on any single metric falls within the bottom 1st percentile of the population.
% \end{itemize}
Based on these process, all problems are partitioned into selected or rejected sets. 
Rejected candidates re-enter the evolutionary cycle for a second round of genetic and crossover operations. 

However, our evolutionary process ends after the second generation. 
This decision is twofold: 
1) Empirical evidence indicates that two generations are sufficient for a significant increase in difficulty.
2) Subsequent evolution can lead to increased verbosity and degraded readability. 
Consequently, our dataset is comprised of first and second-generation problems, although EvolMathEval supports further iterations when required. 
Through this process, EvolMathEval continuously iterates and optimizes, steadily enhancing the difficulty of the dataset.

\section{Experiments}
In this section, we conduct experiments to answer the following five research questions:

\textbf{RQ1:} Does the proposed fitness function serve as an effective and accurate measure of the difficulty of the problem?

\textbf{RQ2:} Does each genetic operator independently contribute to enhancing problem difficulty?

\textbf{RQ3:} Is EvolMathEval capable of creating a continuous evolutionary loop to progressively increase the challenge of the problems?

\textbf{RQ4:} Can EvolMathEval effectively generalize to evolve widely-used public datasets and increase their difficulty?

\textbf{RQ5:} What are the primary failure patterns when LLMs solve these evolved problems, and why do they occur?

\subsection{Experimental Setup}
% To validate the effectiveness and generalization capability of the EvolMathEval framework, we conducted a series of experiments. 
For evaluation, we selected a range of mainstream LLMs, including 
% DeepSeek-v3-0324 \cite{deepseekai2025deepseekv3technicalreport}, deepseek-r1 \cite{deepseekai2025deepseekr1incentivizingreasoningcapability}, gemini-2.5-pro, Kimi-K2 \cite{kimiteam2025kimik2openagentic}, gpt-o3, Doubao-Seed-1.6, GLM-4-flash \cite{glm2024chatglmfamilylargelanguage}, Llama3-70b-8192, Llama3-8b-8192 \cite{grattafiori2024llama3herdmodels}, qwen3-235b-a22b, Qwen3-30b-a3b, and Qwen3-32b \cite{yang2025qwen3technicalreport}{\color{red}{remember to edit}}
DeepSeek-V3 \cite{deepseekai2025deepseekv3technicalreport}, DeepSeek-R1 \cite{deepseekai2025deepseekr1incentivizingreasoningcapability}, Qwen3 \cite{yang2025qwen3technicalreport}, among others. 
The experimental datasets were divided into two categories: 
1) A self-constructed dataset generated from scratch by EvolMathEval; 
2) To test for generalization, we applied EvolMathEval to several widely-used benchmarks.
Specifically, we evolved GSM8K, SVAMP \cite{patel2021nlpmodelsreallyable}, and MAWPS \cite{koncel-kedziorski-etal-2016-mawps} to generate corresponding high-difficulty versions.
% Table~\ref{tab:config_params} shows the key configuration parameters.
Appendix~\ref{appendix:I} shows the key configuration parameters.

% In our experiments, the evolutionary framework was configured with a population size of 300. 
% We name this dataset EvolMath-300. 
% To further expand the data volume for more comprehensive future evaluations, we have also generated an extended set containing 1,000 samples, named EvolMath-1000.
% Each seed problem was generated with 5 variables and 5 equations, with solutions sampled from the integer range [1, 20] and a sparsity of 2 variables per equation. 
% During selection, a fitness score threshold of -0.5 was used to identify and re-evolve problems of insufficient difficulty.

% To ensure the correctness of the problems, we extract the formals of each problem. Then, we check if they contain the seed formals. If the formals are identical, we confirm that the evaluation is successful. Based on a sample of 300 problems, the success rate is 99.3\%.

To ensure the logical integrity of each evolved problem, we performed an automated verification. 
The process involves extracting the mathematical formulas from the final text and confirming that they fully contain the original formulas. 
This check is crucial as it prevents semantic drift that could lead to unsolvable or multi-solution scenarios. 
It was successful for 99.3\% of EvolMath-300, validating the reliability of our generation method.
% Furthermore, we manually reviewed a 30\% sample of the dataset to ensure its readability and coherence.
% Furthermore, a manual review of a 30\% sample confirmed the dataset's readability and coherence.
Furthermore, a manual review of a 30\% sample confirmed the dataset's readability and coherence, and a stability analysis (Appendix~\ref{appendix:F}) confirmed the consistency of our results.

\begin{table*}[t]
\centering
\resizebox{\linewidth}{!}{%
\begin{tabular}{@{}p{3.25cm} p{15cm}@{}}
\toprule
\textbf{Metric} & \textbf{Formula and Description} \\
\midrule
Referee Score & $S_{\text{referee}} \in [0, 10]$: Difficulty rating (0-10) by an expert third-party LLM. \\
\midrule % <-- 这里是修改点

Word Count & $N_{\text{words}}$: The total number of non-punctuation words in the problem text. \\
\cmidrule(lr){1-2}
\begin{tabular}[c]{@{}l@{}}Readability \\ (Flesch-Kincaid)\end{tabular} & $0.39 \left( \frac{N_{\text{words}}}{N_{\text{sentences}}} \right) + 11.8 \left( \frac{N_{\text{syllables}}}{N_{\text{words}}} \right) - 15.59$: Gauges comprehension difficulty by U.S. grade level. \\
\cmidrule(lr){1-2}
\begin{tabular}[c]{@{}l@{}}Syntactic \\ Complexity\end{tabular} & $\max_{s \in \text{doc}} \text{depth}(\text{DepTree}(s))$: Maximum depth of the dependency parse tree across all sentences. \\
\cmidrule(lr){1-2}
\begin{tabular}[c]{@{}l@{}}Word-level \\ Entropy\end{tabular} & $-\sum_{w \in W} p(w) \log_2 p(w)$: Shannon entropy calculated based on the frequency of unique words. \\
\midrule % <-- 这里是修改点

Number of Variables & $|V|$: The count of unique variables identified in the problem. \\
\cmidrule(lr){1-2}
Number of Equations & $|E|$: The count of sentences identified as containing mathematical equations or relations. \\
\cmidrule(lr){1-2}
Noise Ratio & $R_{\text{noise}} = \frac{L_{\text{total}} - L_{\text{signal}}}{L_{\text{total}}}$: The proportion of the text that does not contain mathematical terms. \\
\bottomrule
\end{tabular}%
}
\caption{Overview of the Fitness Function features, grouped into three categories: Heuristic (metric 1), Linguistic (metrics 2-5), and Mathematical (metrics 6-8).}
\label{tab:fitness_features}
\end{table*}

\subsection{Validation of the Fitness Function (RQ1)}
\label{sec:validation_fitness}
This section aims to validate the effectiveness of the proposed fitness function as a measure of problem difficulty. 
To construct an objective, data-driven weighting system, we eschewed a priori manual settings. 
The final weights are determined by the statistical relationship between each feature and the models' problem-solving accuracy.

\textbf{Weight Calculation Method: }
% This method first excludes features with exceptionally high t-test p-values, indicating no significant correlation with model performance. 
% A feature's weight is directly proportional to the absolute value of its Pearson correlation coefficient, while also being modulated by its significance level: 
% a higher p-value results in a more aggressive weight decay. 
% Furthermore, the direction of the weight is adjusted according to our definition of difficulty: 
% features positively correlated with accuracy are assigned negative weights, and vice versa. 
This method first filters out statistically insignificant features (i.e., those with a high p-value). 
The magnitude of a feature's weight is determined by two factors: correlation strength and statistical significance. 
To receive the largest weight, a feature must have both a strong correlation and high significance (a low p-value).
The final step is to set the sign of the weight to reflect difficulty: Features that make problem easier are assigned a negative weight, and vice versa.
Experiments show that although the significance of individual features varies slightly across different models, this method successfully extracts robust difficulty signals common to all models.

% This data-driven weighting mechanism is formalized in the following equation, which computes the comprehensive fitness score ($S$) for a problem from its $M$ extracted features:

\begin{table}[t!]
\centering
\resizebox{\columnwidth}{!}{%
\begin{tabular}{@{}lcc@{}}
\toprule
\textbf{Metric} & \textbf{\textit{p}-value} & \textbf{Weight} \\ \midrule
\multicolumn{3}{l}{\textit{Statistically Significant Metrics (p $<$ 0.05)}} \\
\quad Noise Ratio & 0.009 & +0.19 \\
\quad Lexical Entropy & 0.039 & +0.15 \\
\quad Number of Equations & 0.040 & -0.15 \\
\quad Number of Variables & 0.043 & -0.15 \\ \midrule
\multicolumn{3}{l}{\textit{Moderately Correlated Metrics (p $>$ 0.05, Retained)}} \\
\quad Referee Score & 0.064 & -0.13 \\
\quad Readability Score & 0.130 & -0.10 \\
\quad Word Count & 0.170 & +0.09 \\
\quad Syntactic Complexity & 0.350 & +0.05 \\ \midrule
\multicolumn{3}{l}{\textit{Non-Significant Metrics (p $>$ 0.5, Excluded)}} \\
\quad Semantic Uniqueness & 0.791 & 0.0 \\
\quad Non-linear Relations & 0.902 & 0.0 \\
\bottomrule
\end{tabular}
}
\caption{Data-driven weight calculation for the fitness function, grouped by statistical significance.}
\label{tab:fitness_weights}
\end{table}

\textbf{Weight Calculation Results: }
% This approach yields a nuanced weight vector, as shown in Table~\ref{tab:fitness_weights}. 
Table~\ref{tab:fitness_weights} shows our data-driven weight calculation.
For instance, Noise Ratio received the highest positive difficulty weight (+0.19) due to its high significance (p $<$ 0.01) and strong negative correlation with accuracy. 
Conversely, the weights of less significant features, such as Syntactic Complexity, were substantially attenuated (+0.05). 
% The resulting composite difficulty score demonstrates a statistically significant negative correlation with problem-solving accuracy across multiple representative LLMs (including Qwen and DeepSeek). 
% Therefore, this fitness function is established as a benchmark for difficulty, providing a reliable basis for guiding the evolutionary process toward higher complexity.
Crucially, the final composite score proved to be an effective measure of problem difficulty. 
We found that a higher score consistently correlated with lower problem-solving accuracy across various LLMs, including Qwen and DeepSeek. 
This strong negative correlation validates our fitness function as a reliable guide for the evolutionary process, allowing it to automatically steer the benchmark toward greater complexity.

Notably, Initial investigations explored other difficulty indicators, primarily semantic uniqueness and the count of Non-linear relations. 
However, these features consistently showed non-significant p-values across multiple correlation analyses. 
To construct a more concise and robust fitness function, these features were excluded from calculation.

\subsection{Validation of EvolMathEval's Core Efficacy (RQ2)}
\label{sec:validation_efficacy}

Having established the fitness function as a reliable measure of problem difficulty, this section validates whether EvolMathEval can effectively evolve problems based on this function.
We designed five experiments to measure each operator's impact on problem difficulty. 
The \textbf{Full} setting uses one evolution cycle with all operators. 
Three ablation settings then individually remove the Formulaic Mutations (\textbf{Full w/o FM}), Linguistic Mutations (\textbf{Full w/o LM}), and the Crossover Operator (\textbf{Full w/o CO}). 
A final setting, \textbf{Full (2-gen)}, uses two full cycles to test the effect of secondary evolution.

% Having established the fitness function as a reliable measure of problem difficulty, this section validates whether EvolMathEval can effectively evolve problems based on this metric. To demonstrate that all internal operators of EvolMathEval effectively increase problem difficulty, we designed five rigorous experiments. The configurations are detailed in Table~\ref{tab:ablation_setup}.

% \begin{table}[h!]
% \centering
% \caption{Experimental setup for ablation studies. Operator groups are abbreviated as FD (Formulaic Distractors, Mutations 1-3), LD (Linguistic Distractors, Mutations 4-6), and CO (Crossover Operator). Except for Exp. A, all experiments run for one evolutionary cycle.}
% \label{tab:ablation_setup}
% \begin{tabular}{@{}cc@{}}
% \toprule
% \textbf{Experiment} & \textbf{Operator Configuration} \\ \midrule
% Experiment A & All operators \\
% Experiment B & All operators \\
% Experiment C & Remove FD \\
% Experiment D & Remove LD \\
% Experiment E & Remove CO \\ \bottomrule
% \end{tabular}
% \end{table}

\begin{table*}[t!]
\centering
% 1. 使用标准的小一号字体
% 2. 减小列与列之间的距离 (默认为 6pt)
\resizebox{\textwidth}{!}{%
\begin{tabular}{@{}ccccccccccc@{}}
\toprule
\multicolumn{1}{c}{\textbf{Model}} & \multicolumn{2}{c}{\textbf{Full (2-gen)}} & \multicolumn{2}{c}{\textbf{Full}} & \multicolumn{2}{c}{\textbf{Full w/o FM}} & \multicolumn{2}{c}{\textbf{Full w/o LM}} & \multicolumn{2}{c}{\textbf{Full w/o CO}} \\
\cmidrule(lr){2-3} \cmidrule(lr){4-5} \cmidrule(lr){6-7} \cmidrule(lr){8-9} \cmidrule(lr){10-11}
& Q1 & Final & Q1 & Final & Q1 & Final & Q1 & Final & Q1 & Final \\ \midrule
DeepSeek-V3 & 30.57 & 21.51 & 36.10 & 26.81 & 57.14 & 44.68 & 63.40 & 48.11 & - & 74.07 \\
DeepSeek-R1 & \textbf{87.10} & \textbf{79.03} & \textbf{94.87} & \textbf{87.18} & \textbf{96.88} & \textbf{96.88} & \textbf{100.00} & \textbf{100.00} & - & \textbf{95.56} \\
Doubao-Seed-1.6 & 83.33 & 72.70 & 83.58 & 74.45 & 90.34 & 76.57 & 90.57 & 84.51 & - & 84.67 \\
Gemini-2.5-Pro & 84.03 & 76.47 & 82.00 & 74.00 & 94.00 & 88.00 & 84.00 & 78.00 & - & 90.00 \\
GLM-4-Flash & 26.09 & 2.89 & 22.79 & 3.72 & 43.11 & 7.35 & 21.34 & 2.23 & - & 26.34 \\
GPT-o3 & 76.47 & 65.55 & 88.64 & 68.18 & 84.00 & 80.00 & 82.00 & 76.00 & - & 92.00 \\
Kimi-K2 & 54.87 & 40.71 & 80.00 & 57.50 & 88.00 & 74.00 & 66.00 & 56.00 & - & 71.43 \\
Llama-3-70b & 6.12 & 1.85 & 13.33 & 3.12 & 13.33 & 0.00 & 7.63 & 3.70 & - & 18.40 \\
Llama-3-8b & 4.17 & 0.00 & 7.14 & 8.16 & 4.17 & 0.00 & 4.90 & 2.90 & - & 1.75 \\
Qwen3-235b-a22b & 31.19 & 18.35 & 36.00 & 20.00 & 70.00 & 66.00 & 48.00 & 22.00 & - & 62.00 \\
Qwen3-30b-a3b & 30.10 & 13.09 & 40.67 & 18.39 & 76.09 & 55.56 & 47.67 & 25.00 & - & 47.28 \\
Qwen3-32b & 25.68 & 6.73 & 37.12 & 16.44 & 72.24 & 57.05 & 41.33 & 19.33 & - & 48.66 \\ 
\bottomrule
\end{tabular}
}
\caption{Model accuracy (\%) across five different experimental settings. Q1 and Final refer to the accuracy on the first and final sub-problems, respectively, of compound questions generated by the crossover operator.}
\label{tab:ablation_results}
\end{table*}

\textbf{Validation of the Operators' Efficacy: }
% The results of all five experimental versions are presented in Table~\ref{tab:ablation_results}.
% The crossover operator demonstrated the most significant impact on problem difficulty. 
% The accuracy of most models in the Full w/o CO setting was significantly higher than in any other evolved version.
% For example, DeepSeek-V3 achieved an accuracy of 74.07\%. 
% However, upon introducing the crossover operator (i.e., in the Full setting), its final accuracy plummeted to 26.81\%. 
% This indicates that crossover operator is the primary factor in increasing problem difficulty.
The results in Table~\ref{tab:ablation_results} confirm the efficacy of each operator. 
The Crossover Operator has the most significant impact on problem difficulty.
For example, DeepSeek-V3's accuracy drops from 74.07\% (in Full w/o CO) to 26.81\% (in Full).

The formulaic mutations also proved to be a crucial component. 
% With their inclusion, model accuracy drops noticeably. 
For instance, the final accuracy of Qwen3-30b-a3b rebounded from 18.39\% (in Full) to 55.56\% in the Full w/o FM. 
This proves that formula-level modifications, are also important for increasing problem complexity.

Additionally, the secondary evolution mechanism pushed the difficulty ceiling even higher. 
Across all settings, almost every model achieved its lowest accuracy in Full (2-gen). 
Taking DeepSeek-V3 as an example, its accuracy decreased further from 26.81\% in Full to 21.51\% in Full (2-gen). 
This demonstrates that the evolutionary process can further enhance the difficulty on top of the synergistic effects of all operators.

\textbf{Impact of Problem Structure on Reasoning: }
We observed that simply altering the problem structure impacts reasoning ability. 
For instance, DeepSeek-V3's accuracy on problems of Full w/o CO was 74.07\%, but when the same type of problem was presented as Full, its accuracy plummeted to 36.1\% (Q1 accuracy). 
% This sharp decline suggests that adding a dependent follow-up task imposes significant comprehension pressure, which we find that it often causing models to erroneously ignore the first sub-problem entirely.
This sharp decline suggests that a dependent follow-up task creates significant comprehension pressure. 
Our analysis of the failure cases revealed that this pressure often caused the systems to erroneously ignore the first sub-problem altogether.
% Additionally, we observed a significant performance discrepancy between the final accuracy in Full w/o CO and the Q1 accuracy in Full. Theoretically, since the problems in both scenarios have been processed by all mutation operators, their intrinsic complexity should be highly similar. Thus, one would expect the Final Accuracy of Full w/o CO to be roughly equivalent to the Q1 Accuracy of Full. \textbf{However, the results in Table~\ref{tab:ablation_results} reveal the opposite.} For DeepSeek-V3, the Q1 accuracy in Full was only 36.1\%, far below the 74.07\% achieved on similar problems in Full w/o CO. This implies that a mere change in problem presentation structure can significantly impact a model's reasoning ability. Our analysis revealed a notable increase in errors where the model ``ignores the first question and directly solves the second". This suggests that even adding a single dependent follow-up task is enough to impose significant comprehension pressure and degrade accuracy. A detailed analysis is provided in Section~\ref{sec:badcase_analysis}.

\subsection{Validation of Dynamic Evolution Capability (RQ3)}
After validating the effectiveness of the fitness function (Section~\ref{sec:validation_fitness}) and the functionality of the core genetic operators (Section~\ref{sec:validation_efficacy}), the primary objective of this section is to demonstrate that these modules can work in concert to enable the ``evolvability" of the EvolMathEval framework. 
% To this end, we designed an experiment to verify its dynamic evolutionary capabilities.

% We first performed one round of evolution and used the fitness function to partition the generated problems into "Qualified" and "Unqualified" subsets. Subsequently, we applied a second round of evolution exclusively to the "Unqualified" subset, creating a "SecondaryEvolution" population. Finally, we evaluated LLM accuracy on these three datasets.
We first performed one round of evolution. 
The resulting problems were then partitioned by the fitness function into two subsets: 
An \textbf{Easy} set (`Unqualified' problems) and a \textbf{Hard} set (`Qualified' problems). 
Subsequently, we applied a second round of evolution exclusively to the Easy set to create the Evolved Easy set. 
Finally, we evaluated the accuracy of LLM in these three datasets.
The results can be seen in Table~\ref{tab:final_accuracy_summary}.

\begin{table}[h!]
\centering
\resizebox{\columnwidth}{!}{%
\begin{tabular}{l c c c}
\toprule
\textbf{Model} & \textbf{Easy} & \textbf{Hard} & \textbf{Evolved Easy} \\
\midrule
DeepSeek-V3    & 31.25 & 23.46 & 10.84 \\
DeepSeek-R1         & 89.66 & 88.76 & 80.95 \\
Doubao-Seed-1.6     & 78.82 & 63.77 & 49.18 \\
Gemini-2.5-Pro      & 90.00 & 75.00 & 76.81 \\
GLM-4-Flash         & 4.84  & 2.99  & \textbf{0.00}  \\
GPT-o3              & 71.91 & 65.96 & 37.93 \\
Kimi-K2             & 54.29 & 42.50 & \textbf{0.00}  \\
Qwen3-235b-a22b     & 30.91 & 25.00 & 10.14 \\
Qwen3-30b-a3b       & 12.90 & 10.23 & 3.33  \\
Qwen3-32b           & 12.09 & 7.55  & 4.60  \\
\bottomrule
\end{tabular}
}
\caption{
% Final Accuracy (\%) of Models. After the evolution cycle, the problems are partitioned by the fitness function into the Easy and Hard sets. The Evolved Easy set is generated by applying a second evolution cycle to the Easy set.
Model accuracy (\%) on three problem sets: the "Easy" and "Hard" sets partitioned after one evolution, and the "Evolved Easy" set created by re-evolving the "Easy" set.
}
\label{tab:final_accuracy_summary}
\end{table}

% \textbf{Effectiveness of Fitness Function: }As shown in Table~\ref{tab:final_accuracy_summary}, the results clearly demonstrate the framework's dynamic evolutionary efficacy. Model accuracy on the "Qualified" set is significantly lower than on the "Unqualified" set. For example, Doubao-Seed-1.6's final accuracy dropped from 78.82\% on the "Unqualified" set to 63.77\% on the "Qualified" set. This confirms that the fitness function can effectively screen for less difficult problems.

\textbf{Effectiveness of Fitness Function: }
As shown in Table~\ref{tab:final_accuracy_summary}, 
% the effectiveness of the fitness function is demonstrated by comparing the Easy and Hard columns. 
for all models, accuracy on the Hard set is lower than on the Easy set.
For example, Doubao-Seed-1.6's accuracy dropped from 78.82\% to 63.77\%. 
This confirms that the fitness function can effectively filter problems by difficulty.

% \textbf{Effectiveness of Secondary Evolution: }Most critically, secondary evolution significantly increases problem complexity. Compared to the Easy set, all models experienced a precipitous drop in accuracy on the Evolved Easy set. For instance, Doubao-Seed-1.6's accuracy plummeted from 78.82\% to 49.18\%. This result proves that the genetic operators can systematically enhance problem difficulty. Moreover, the difficulty of the Evolved Easy set (e.g., 10.84\% accuracy for deepseek-v3-0324) often surpassed that of the initial Easy set (23.46\%), revealing the framework's potential to continuously generate more formidable challenges.

% Interestingly, Kimi K2 achieves an accuracy of 54.29\% on the Easy set, but its accuracy plummets to 0.00\% on the Evolved Easy set. This is possibly due to the 'Budget Control' mechanism used during its reinforcement learning phase, which enhances the model's token efficiency \citep{kimiteam2025kimik2openagentic}. However, when performing long-chain reasoning, the model may be constrained by its 'reasoning budget,' preventing it from conducting an in-depth analysis during inference, which ultimately leads to the failure of the entire task.

\textbf{Effectiveness of Secondary Evolution: }
Most critically, secondary evolution significantly increases problem complexity, Kimi-K2 is an example. 
% This model, which achieves a respectable 54.29\% accuracy on the Easy set, sees its performance plummet to 0.00\% on the Evolved Easy set.
The model's performance plummeted from 54.29\% on the Easy set to 0.00\% on the Evolved Easy set.
This drastic failure is possibly due to the \textbf{`Budget Control'} mechanism from its reinforcement learning phase, which is designed to enhance token efficiency \citep{kimiteam2025kimik2openagentic}. 
This `reasoning budget' may be effective for simpler tasks. 
However, it may also constrain the model from performing the in-depth analysis required for long-chain reasoning, leading to complete task failure. 

Furthermore, the secondary evolution is so pronounced that the Evolved Easy set often becomes more challenging than the initial Hard set. 
For example, DeepSeek-V3's accuracy on the Evolved Easy set (10.84\%) was less than half its accuracy on the Hard set (23.46\%), demonstrating EvolMathEval's capacity to create more challenging problems.

In summary, this experiment fully validates the dynamic evolutionary loop of EvolMathEval: 
It can accurately identify low-difficulty problems, effectively enhance them using genetic operators, and ultimately achieve stable and iterative growth in benchmark difficulty.
% To contextualize the difficulty of our generated benchmark, a direct comparison against other challenging datasets is presented in Appendix~\ref{appendix:G}.
To further measure the capability of EvolMathEval, we compared EvolMath with other challenging benchmarks like AMWPG \cite{Xie2024AdversarialMW} and GSM-PLUS \cite{li-etal-2024-gsm}.
The results are presented in Appendix~\ref{appendix:G}.

\subsection{Validation of Generalization Capability (RQ4)}
\label{Validation_of_Generalization_Capability}
Building on the previous validation of the EvolMathEval framework's effectiveness, this section evaluates its generalization capability on external datasets. 
We selected three widely-used public mathematical reasoning datasets: GSM8K, SVAMP, and MAWPS. 
The experimental process involved first using an LLM to extract the core mathematical formulas from these datasets. 
Then applying EvolMathEval to these formulas to generate corresponding high-difficulty versions.

\begin{table}[t!]
\centering
% \caption{Relative Accuracy Drop (\%) on Public Benchmarks After Evolution. Values represent the percentage decrease relative to the original score, calculated as (Original - Evolved) / Original. The final row shows the average relative drop.}
\resizebox{\columnwidth}{!}{%
\begin{tabular}{l c c c}
\toprule
\textbf{Model} & \textbf{GSM8K} & \textbf{SVAMP} & \textbf{MAWPS} \\
\midrule
DeepSeek-V3        & 61.69        &        45.60        &        54.49 \\
DeepSeek-R1        & 47.57        &        25.22        &        35.46 \\
Doubao-Seed-1.6    & 42.85        &        26.43        &        32.96 \\
Gemini-2.5-Pro     & 35.42        &        26.61        &        30.43 \\
GLM-4-Flash        & \textbf{94.68}        &        \textbf{80.45}        &        \textbf{90.89} \\
GPT-o3             & 38.14        &        23.47        &        33.57 \\
Kimi-K2            & 43.54        &        35.56        &        33.49 \\
Llama-3-70b    & 67.77        &        51.52        &        60.52 \\
Llama-3-8b     & 86.37        &        65.94        &        86.40 \\
Qwen3-235b-a22b    & 47.92        &        50.00        &        40.00 \\
Qwen3-30b-a3b      & 54.47        &        32.48        &        45.18 \\
Qwen3-32b          & 54.18        &        34.73        &        42.00 \\
\midrule
\textbf{Average Rel. Drop} & \textbf{56.22} & \textbf{41.50} & \textbf{48.78} \\
\textbf{Average Evolved Acc.} & \textbf{41.93} & \textbf{56.03} & \textbf{47.94} \\
\bottomrule
\end{tabular}
}
\caption{
% Performance on Evolved Public Benchmarks. The main data shows the relative accuracy drop (\%) for each model. The final two rows show the average relative drop and the average absolute accuracy (\%) on the evolved datasets, respectively.
Relative accuracy drop (\%) on evolved public benchmarks. The final two rows summarize the average drop and the average absolute accuracy.
}
\label{tab:accuracy_drop}
\end{table}

\begin{table*}[h!]
\centering
\resizebox{\textwidth}{!}{%
\begin{tabular}{@{}p{3.35cm} p{6.5cm} p{7.1cm}@{}}
\toprule
\textbf{Error Pattern} & \textbf{Description} & \textbf{Examples of LLM Failures} \\ \midrule

\textbf{Pseudo Aha Moment} & Adopts ambiguous, logically flawed ``reasoning shortcuts" instead of rigorous multi-step logic. A example is shown in Figure~\ref{fig:pseudo_eureka_example}.& 
1) Incorrectly treats an approximate condition (e.g., $A \approx B$) as a precise equation.
\\
\cmidrule(l){1-3}

\textbf{Variable Mapping Errors} & Fails to accurately map natural language descriptions to algebraic symbols. & 
1) Confusing variables with similar definitions.
% ('number of apples' vs. 'weight of apples').
    
2) Hallucinating a simple formula instead of using the complex one provided in the text.
\\
\cmidrule(l){1-3}

\textbf{Structural Failures in Reasoning Chains} & 
Fails to maintain logical coherence in multi-step problems.
% created by the Crossover Operator. 
& 
1) Ignoring the first sub-problem and attempting to solve the second part directly.

2) Correctly solving the first step but failing to pass the result to the subsequent step.
\\ \bottomrule
\end{tabular}%
}
\caption{Taxonomy of LLM Reasoning Failures on Evolved Mathematical Problems.}
\label{tab:error_patterns_examples}
\end{table*}

\textbf{Enhancement of Problem Difficulty:}
The results (Table~\ref{tab:accuracy_drop}) reveal the powerful generalization capability of the EvolMathEval(see Appendix~\ref{appendix:D} for the full absolute accuracy scores). 
After applying it, all tested models showed a significant decline in accuracy across all datasets. 
For example, on the GSM8K dataset, models experienced a significant performance reduction, with GLM-4-Flash's accuracy decreasing by 94.68\% from its original score. 
It demonstrates that EvolMathEval can effectively increase the difficulty of existing benchmarks.

% Beyond increasing absolute difficulty, a more significant value of our framework lies in its ability to enhance the differentiability of model capabilities. On the original GSM8K dataset, the accuracies of several top models were highly concentrated in the narrow 96\%-97.5\% range, making performance differences difficult to discern. However, after evolution with EvolMathEval, model performances were effectively separated into a clear gradient: from the relatively robust Doubao-Seed-1.6 (55.15\%) to the nearly collapsing glm-4-flash (4.51\%). This directly uncovers true capability differences that were previously masked by high scores on simpler benchmarks.

\textbf{Enhancement of Model Discriminative Power:}
% Beyond demonstrating a universal increase in difficulty, the results also highlight EvolMathEval's ability to \textbf{enhance model differentiability}. 
While top models often cluster together with near-perfect scores on the original benchmarks, the evolved versions effectively separate them, revealing a wide spectrum of performance.
On GSM8K, for example, the relative accuracy drop ranged from 35.42\% for Gemini-2.5-Pro to 94.68\% for GLM-4-Flash.
This wide variance uncovers true capability differences previously masked by score saturation.

% \textbf{Synergistic Difficulty Enhancement: }Additionally, the framework's difficulty enhancement exhibits a systematic, synergistic effect that preserves the inherent difficulty hierarchy of the original datasets. After evolution, all models performed significantly worse on GSM8K than on SVAMP and MAWPS. This correlates directly with the initial complexity of the datasets: the average reasoning chain lengths for GSM8K, SVAMP, and MAWPS are 3.905, 2.235, and 1.965, respectively. Our framework's evolutionary operators produce a magnified synergistic effect with longer initial reasoning chains, causing problems with longer original chains to experience a more pronounced increase in difficulty after evolution.

% Additionally, the framework's difficulty enhancement exhibits a systematic, synergistic effect that preserves the inherent difficulty hierarchy of the original datasets. 

Additionally, the framework's enhancement effect is systematic: It makes harder datasets more challenging while preserving their original difficulty ranking.
As shown in the final row of Table~\ref{tab:accuracy_drop}, the average relative accuracy drop was most severe on GSM8K (56.22\%), followed by MAWPS (48.78\%) and SVAMP (41.50\%). 
This ranking directly correlates with each dataset's initial complexity, as measured by their average reasoning chain lengths (3.905, 2.235, and 1.965, respectively).
This is because our evolutionary operators amplify the difficulty of problems more effectively when they already have longer reasoning chains.

% To 验证 the 挑战性 of EvolMath, we compare EvolMath with AMWPG \cite{Xie2024AdversarialMW} and GSM-PLUS \cite{Li2024GSMPlusAC}. The result can be seen in 附录\label{appendix:G}

% Xie2024AdversarialMW
% 如果上面部分删除的话，这里也同样删除后面这句话

\subsection{Badcase Analysis (RQ5)}
\label{sec:badcase_analysis}

% Having demonstrated the capabilities of the EvolMathEval framework, this section investigates the specific reasons for LLM failures on EvolMath by analyzing incorrect solutions and quantitatively validating key patterns discovered.

This section analyzes why LLMs fail on EvolMath. 
We identify and quantify the primary error patterns by analyzing incorrect solutions, which can be seen in Table~\ref{tab:error_patterns_examples}.
Appendix~\ref{appendix:E} provides a detailed, step-by-step analysis of the ``Pseudo Aha Moment''.

\begin{figure}[h!]
    \centering
    \includegraphics[width=\columnwidth]{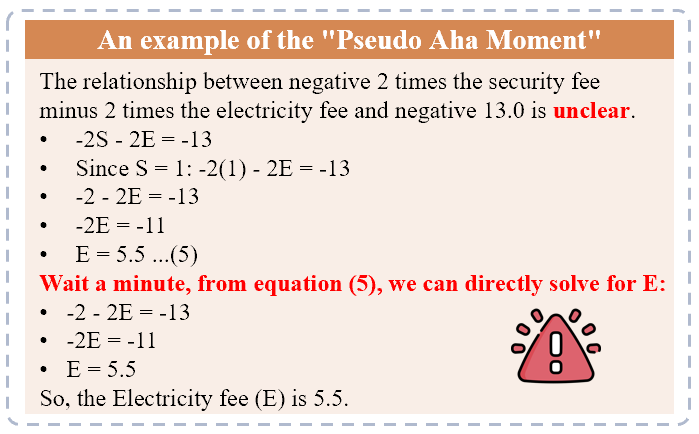}
    \caption{An example of the ``Pseudo Aha Moment" phenomenon, where the model follows a incorrect shortcut introduced by the mutation operator.}
    % 这里后续改一下标题
    \label{fig:pseudo_eureka_example}
\end{figure}

% \paragraph{Analysis of Typical Error Patterns:}
% Through a systematic analysis of incorrect solutions to the evolved problems, we identified three primary error patterns. These patterns, their descriptions, and their primary causes are summarized in Table~\ref{tab:error_patterns}.

\begin{figure}[h!]
    \centering
    \includegraphics[width=\columnwidth]{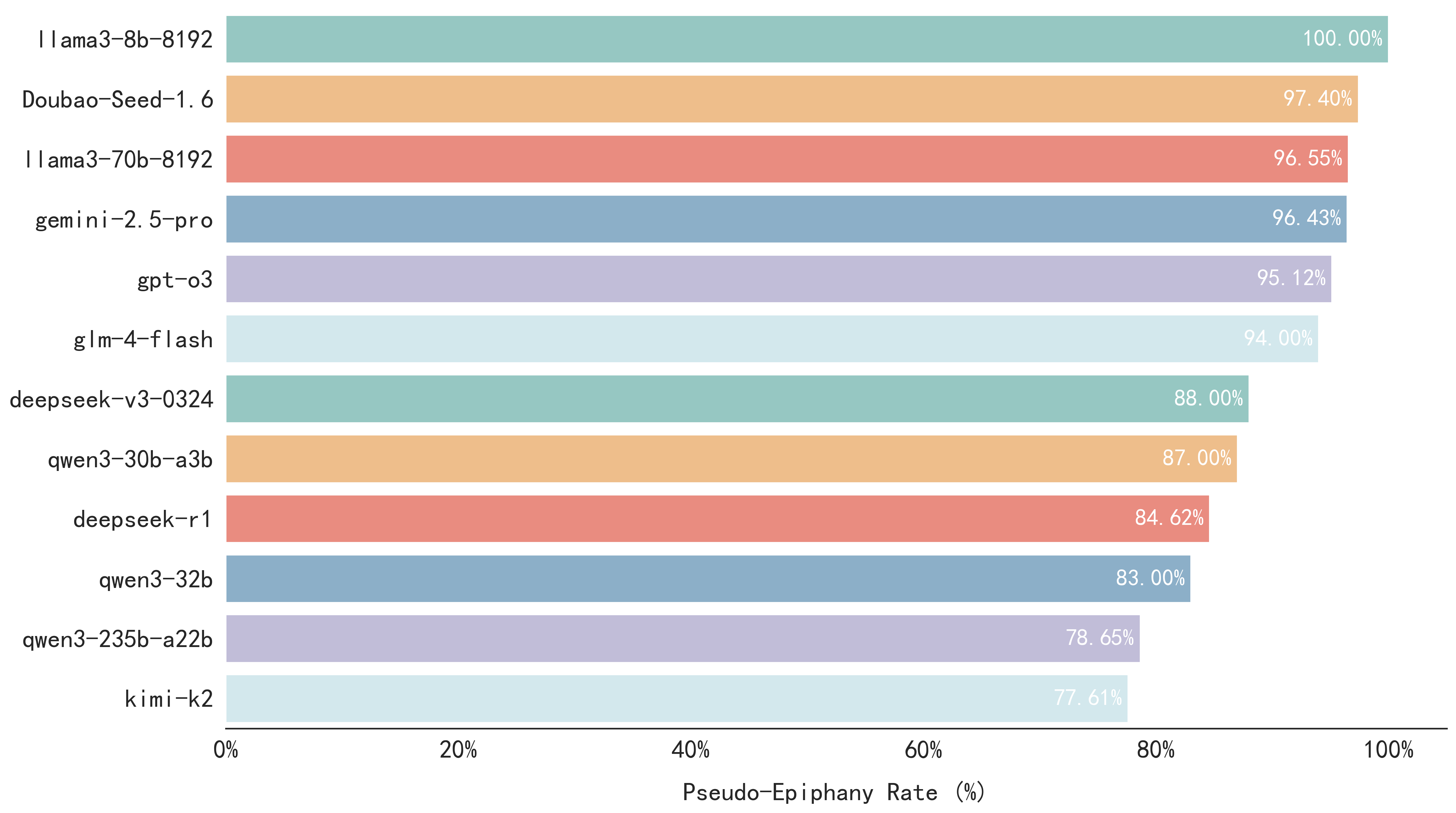}
    \caption{Incidence of "Pseudo Aha Moment" in Failed Solutions to Problems with Approximate Substitution.}
    % 这里后续改一下标题
    \label{fig:pseudo_eureka_rate}
\end{figure}

Of these patterns, the \textbf{Pseudo Aha Moment} was the most prevalent. 
To quantify this, we collected all problems generated by the ``approximate substitution" operator and analyzed the incorrect answers. 
To ensure the objectivity of this analysis, two annotators independently classified errors on a random subset of 50 instances.
The resulting Cohen's Kappa was 0.92, indicating substantial agreement.
As shown in Figure~\ref{fig:pseudo_eureka_rate}, the Rate of this phenomenon is extremely high, even reaching 100\% in some models. 
% This result powerfully reveals an inherent limitation of current LLMs: a tendency to rely on heuristic shortcuts rather than rigorous logic when faced with complex reasoning. 
% It also serves as inverse proof that EvolMathEval, by introducing such challenges, can effectively detect and amplify the deep-seated reasoning flaws of these models.
This result reveals a key limitation of current LLMs: they tend to rely on simple shortcuts instead of rigorous logic for complex reasoning. 
It also proves that EvolMathEval is effective at detecting and amplifying these deep-seated flaws.

\section{Conclusion}
% 6. 结论
% This paper introduces EvolMathEval, an automated mathematical benchmark evolution framework designed to address the issues of score saturation and data contamination. By simulating an evolutionary process, the framework can dynamically generate novel, high-difficulty mathematical problems, thereby ensuring the long-term validity and challenge of the evaluation.

% Experiments demonstrate that the framework not only generates high-difficulty datasets from scratch that reduce the accuracy of top-tier models (e.g., DeepSeek-R1) to 79\%, but it also significantly degrades performance on existing benchmarks like GSM8K through evolution. This results in an average performance drop of 48\% and a decline of up to 95\% for some models. Problems that undergo a second round of evolution can cause the accuracy of certain models, such as Kimi-K2, to plummet to zero. This performance decline reveals true capability differences that were previously masked by high scores on the original benchmarks.

% More importantly, this work is the first to uncover and quantify a weakness in large models, which we term the ``Pseudo Aha Moment". When faced with the complex challenges we introduce, models tend to adopt logically flawed ``shortcuts" instead of rigorous reasoning. Our quantitative analysis shows that this is prevalent in the models' incorrect answers, with an incidence rate ranging from a minimum of 77\% to as high as 100\% for some models.

In this work, we introduced EvolMathEval, an automated framework that applies evolutionary testing to dynamically generate and evolve novel and challenging problems.

Our experiments confirm the framework's efficacy. 
EvolMathEval is effective in two ways: 
It can create challenging problems from scratch and significantly harden existing datasets like GSM8K. 
This process is highly effective, causing model accuracies to drop by 48\% on average. In some cases, model performance collapsed to zero.

% More importantly, this process uncovers a core vulnerability in current LLMs, which we term the "Pseudo Aha Moment"—a tendency to favor flawed, heuristic shortcuts over rigorous reasoning. 
% Our analysis reveals this is a dominant failure mode, responsible for 77\% to 100\% of errors on targeted problems, highlighting a deep-seated weakness in their reasoning capabilities.

More importantly, this process uncovers a core weakness in current LLMs, which we term the "Pseudo Aha Moment." 
This is a tendency for LLMs to favor flawed, simple shortcuts over rigorous, multi-step reasoning.
Our analysis confirms this is a dominant failure mode on targeted problems, it was responsible for 77\% to 100\% of all errors, highlighting a deep-seated weakness in LLMs' reasoning capabilities.

\bibliography{custom}

\appendix
% \section{Appendix}

% This appendix provides supplementary materials and detailed explanations for the main paper. 

% Appendix A elaborates on the mathematical design of the ``Necessity Check," a critical validation step that ensures the effectiveness of the Crossover Operator within our evolutionary framework. 

% Appendix B offers a comprehensive overview of the entire EvolMathEval problem generation pipeline, systematically detailing all stages from seed problem initialization, multi-stage mutation, and problem translation and polishing, to the final evolution and evaluation, supplemented with corresponding algorithms and figures. 

% Appendix C presents other prompts used in our experiments, including those for difficulty scoring by a ``judge" large language model and for analyzing specific errors like the "Pseudo Aha Moment." 

% Appendix D presents detailed experimental data in tabular form, showing a performance comparison of several mainstream large language models on public benchmarks before and after evolution by our framework. 

% Appendix E provides an in-depth analysis of a ``Pseudo Aha Moment" failure case, dissecting a complete example to illustrate how a model can be misled by a deceptive shortcut, leading to an incorrect reasoning process.

\section{Details of the problem generation of EvolMathEval}
\label{DetailsofEvolMathEval}

This section outlines the complete problem generation process of EvolMathEval. 

The process begins with Seed Problem Initialization, where the system generates initial systems of linear equations, as detailed in Algorithm \ref{alg:Seed_Problem_Initialization}.

\begin{algorithm*}[htbp]
\caption{Seed Problem Initialization}
\label{alg:Seed_Problem_Initialization}
\begin{algorithmic}[1]
    \State $DataSet \gets \emptyset$
    \For{$i \gets 1$ to $N_{prob}$}
        \Loop
            \State Generate a random integer solution vector $\vec{s} \in \mathbb{Z}^n$.
            \State Construct a sparse $n \times n$ coefficient matrix $A$ such that:
            \State \quad a) The first $n-1$ rows only have non-zero entries in the first $n-1$ columns.
            \State \quad b) The last row has non-zero entries for variable $v_{n-1}$ and at least one other variable.
            
            \State Let $A_{sub}$ be the top-left $(n-1) \times (n-1)$ submatrix of $A$.
            \If{$\text{rank}(A_{sub}) < n-1$}
                \State \textbf{continue}
            \EndIf
            
            \State Let $I_{vars}$ be the column indices of variables (other than $v_{n-1}$) in the last equation.
            \State Let $A_{check}$ be the submatrix of $A$ with rows $1, \dots, n-2$ and columns from $I_{vars}$.
            \If{$\text{rank}(A_{check}) = |I_{vars}|$}
                \State \textbf{continue}
            \EndIf

            % \State \Comment{\textbf{Step 3: Finalize and store the valid problem}}
            \State $\vec{b} \gets A \vec{s}$ 
            \State Format the system $A\vec{x}=\vec{b}$ into a `prompt` string.
            \State Format the solution $\vec{s}$ into an `answer` string.
            \State Add $\{ \text{id}: i, \text{prompt}: \text{prompt}, \text{answer}: \text{answer} \}$ to $DataSet$.
            \State \textbf{break}
        \EndLoop
    \EndFor
    \State Save $DataSet$ to a JSON file.
    \State \Return $DataSet$.
\end{algorithmic}
\end{algorithm*}

\begin{figure*}[htbp]
    \centering
    \includegraphics[width=\textwidth]{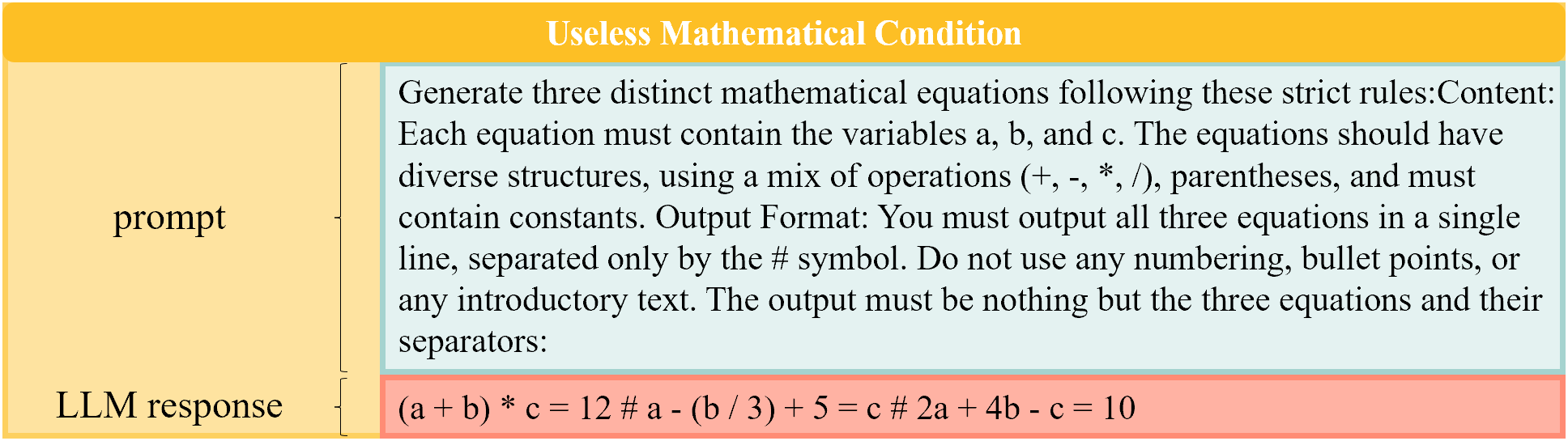}
    \caption{Useless Mathematical Condition}
    \label{fig:Useless_Mathematical_Condition}
\end{figure*}

\begin{figure*}[htbp]
    \centering
    \includegraphics[width=\textwidth]{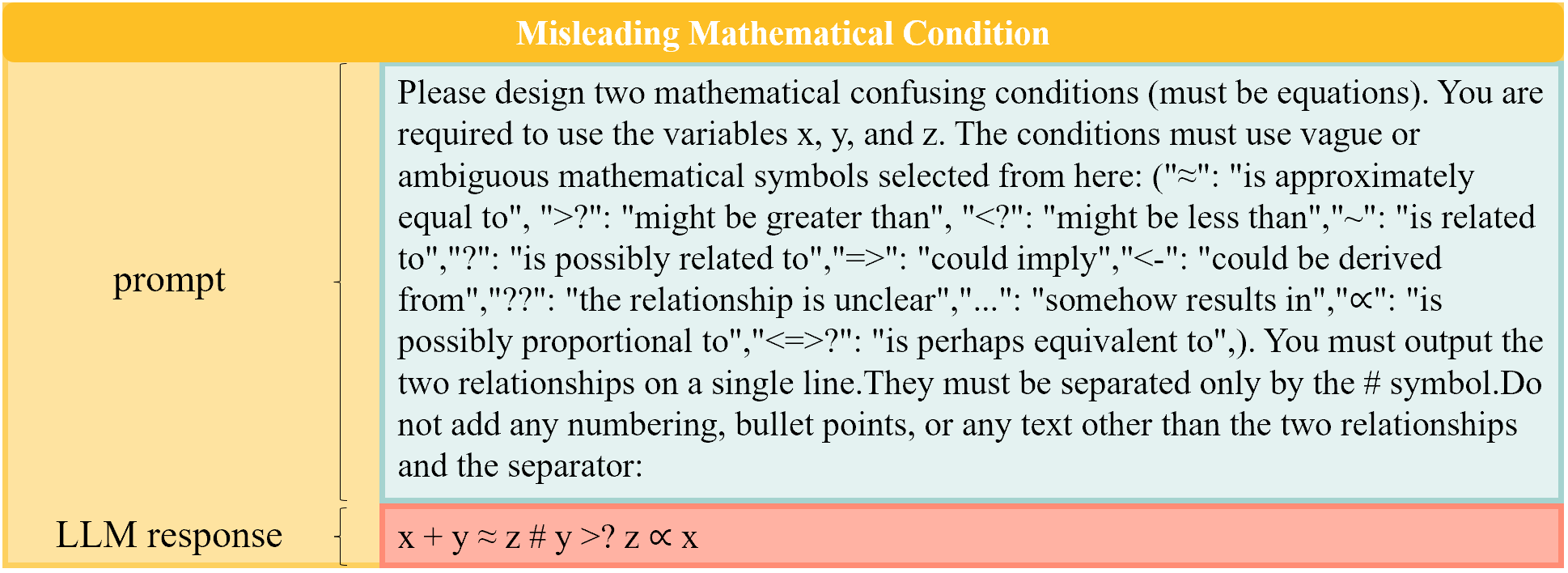}
    \caption{Misleading Mathematical Condition}
    \label{fig:Misleading_Mathematical_Condition}
\end{figure*}

The problems then enter the genetic operator stage, which consists of several core steps:
\begin{itemize}
    \item \textbf{Approximate Replacement}: 
    Generates shortcut formulas that offer alternative solution paths (see Algorithm~\ref{alg:Approximate_Replacement}).
    \item \textbf{Useless Mathematical Condition}: 
    Introduces redundant formulas that do not affect the problem-solving process (see Figure~\ref{fig:Useless_Mathematical_Condition}).
    \item \textbf{Misleading Mathematical Condition}: 
    Creates mathematical conditions designed to confuse the solver's thought process (see Figure~\ref{fig:Misleading_Mathematical_Condition}).
\end{itemize}

\begin{algorithm*}[htbp]
\caption{Approximate Replacement}
\label{alg:Approximate_Replacement}
\begin{algorithmic}[1]
    \State $DataSet \gets$ Read JSON data from $inputFile$.
    \State Define $ConfusionMap$ of ambiguous symbols.

    \For{each $Item$ in $DataSet$}
        \State Let $InitialEquations$ be the list of equations in the item's prompt.
        \If{a directly-assignable variable ($v$) can be found in $InitialEquations$ AND the ``Templated Reconstruction" strategy using $v$ succeeds}
            \State $confused\_eq \gets$ The result of the successful reconstruction.
        \Else
            \State $confused\_eq \gets$ The result of applying ``Random Interference" to the last equation.
        \EndIf
        
        \State Append the generated $confused\_eq$ to the item's prompt.
    \EndFor
    
    \State Save the modified $DataSet$ to $outputFile$.
\end{algorithmic}
\end{algorithm*}

The \textbf{Problem Translation Stage} follows, where the mathematical formulas are initially converted into natural language using a rule-based substitution method (see Algorithm~\ref{alg:Problem_Translation}).

\begin{algorithm*}[htbp]
\caption{Problem Translation}
\label{alg:Problem_Translation}
\begin{algorithmic}[1]
    \State $DataSet \gets$ Read JSON data from $inputFile$.
    \State $EntityLibrary \gets$ Load a predefined library of real-world entities.

    \For{each $Problem$ in $DataSet$}
        \State Create a random `mapping' by assigning a unique real-world entity to each mathematical variable found in the $Problem$.
        \State Using this `mapping', translate every formula into a full natural language sentence.
        \State Replace the original prompt with the generated sentences and store the entity `mapping' information within the $Problem$.
    \EndFor
    
    \State Save the fully translated $DataSet$ to $outputFile$.
\end{algorithmic}
\end{algorithm*}

\begin{figure*}[htbp]
    \centering
    \includegraphics[width=\textwidth]{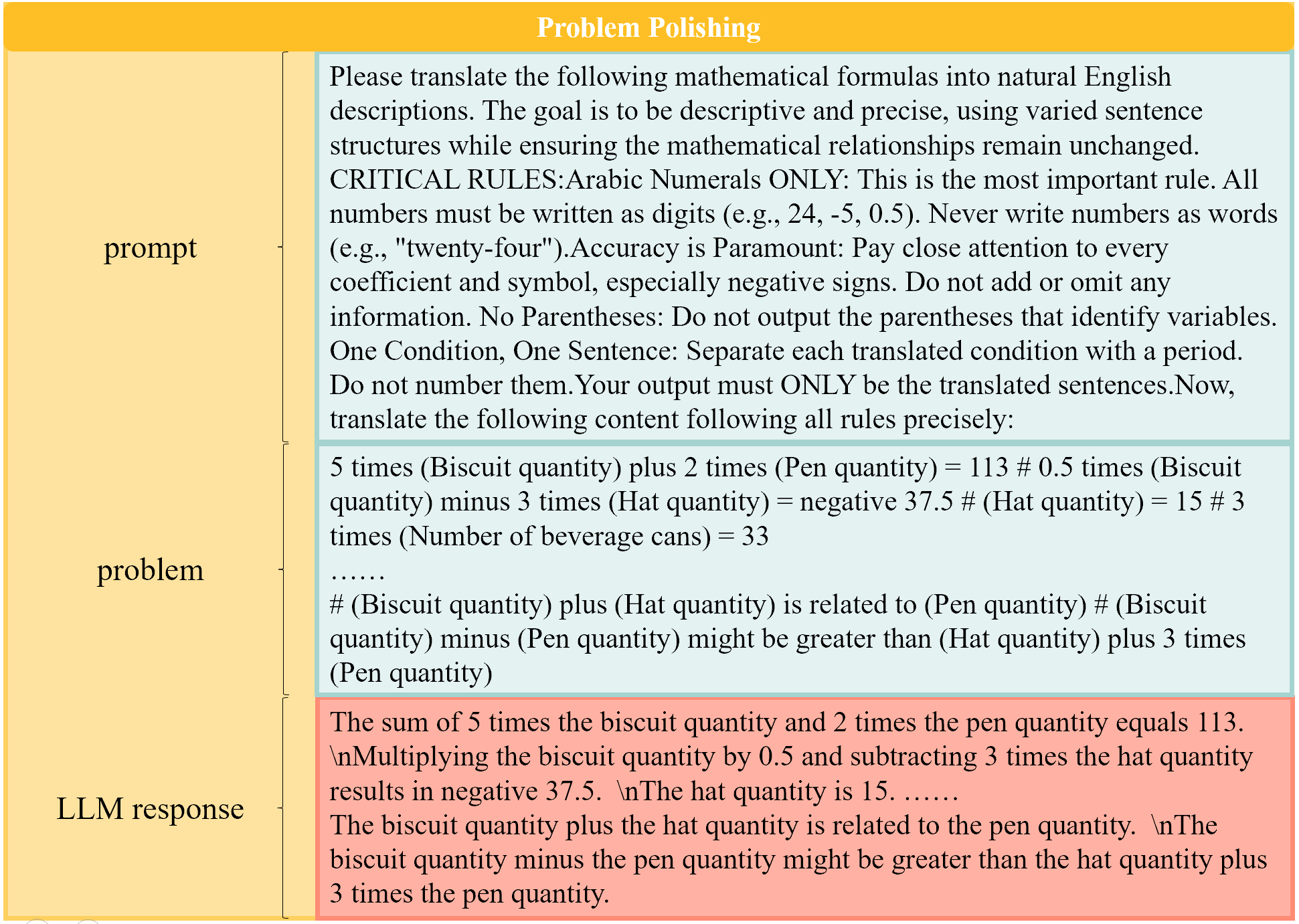}
    \caption{Problem Polishing}
    \label{fig:Problem_Polishing}
\end{figure*}

Subsequently, the \textbf{Problem Polishing Stage} utilizes a large language model to refine the preliminary natural language problems, enhancing their fluency and readability (as shown in Figure~\ref{fig:Problem_Polishing}).

\begin{figure*}[htbp]
    \centering
    \includegraphics[width=\textwidth]{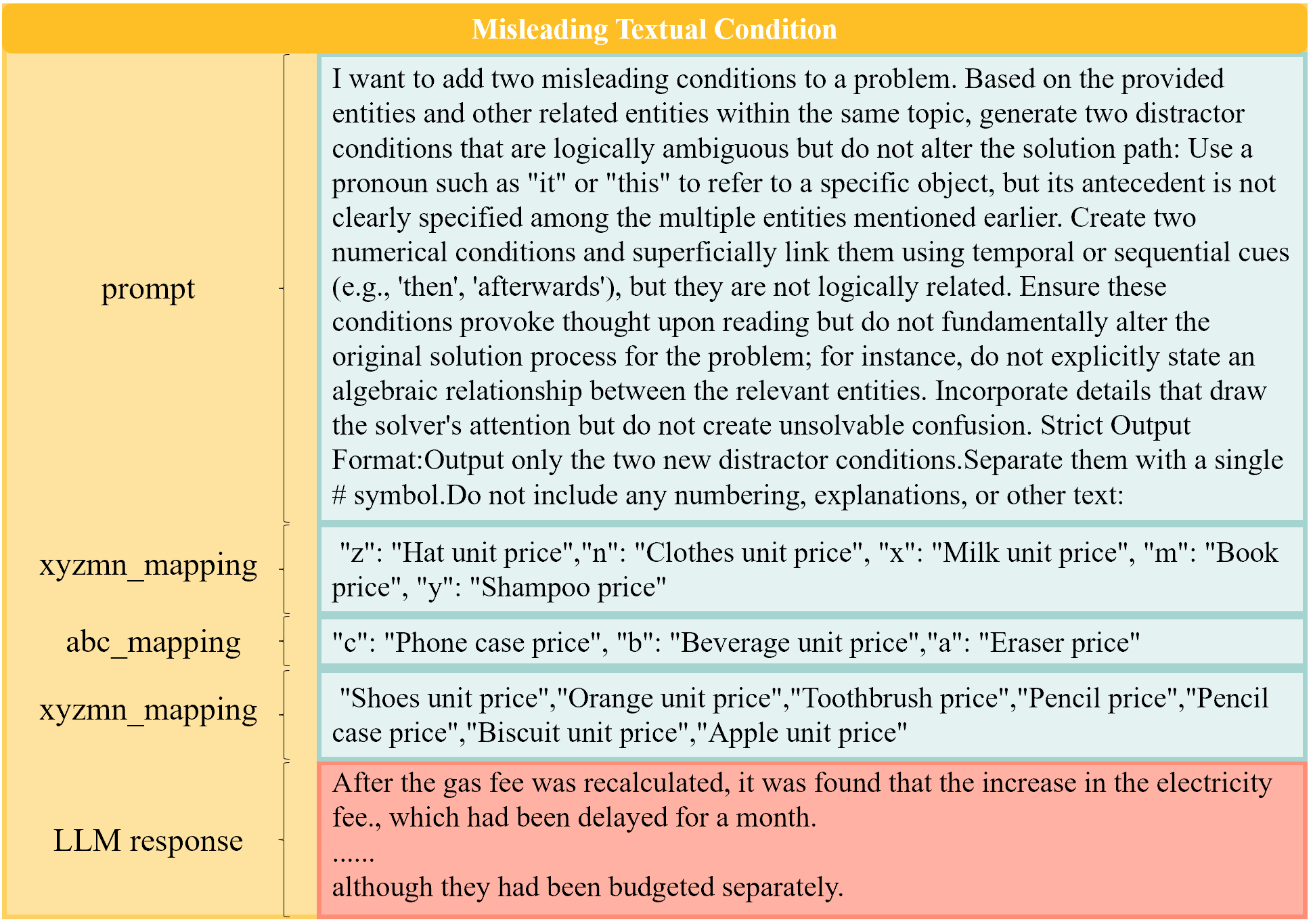}
    \caption{Misleading Textual Condition}
    \label{fig:Misleading_Textual_Condition}
\end{figure*}

\begin{figure*}[htbp]
    \centering
    \includegraphics[width=\textwidth]{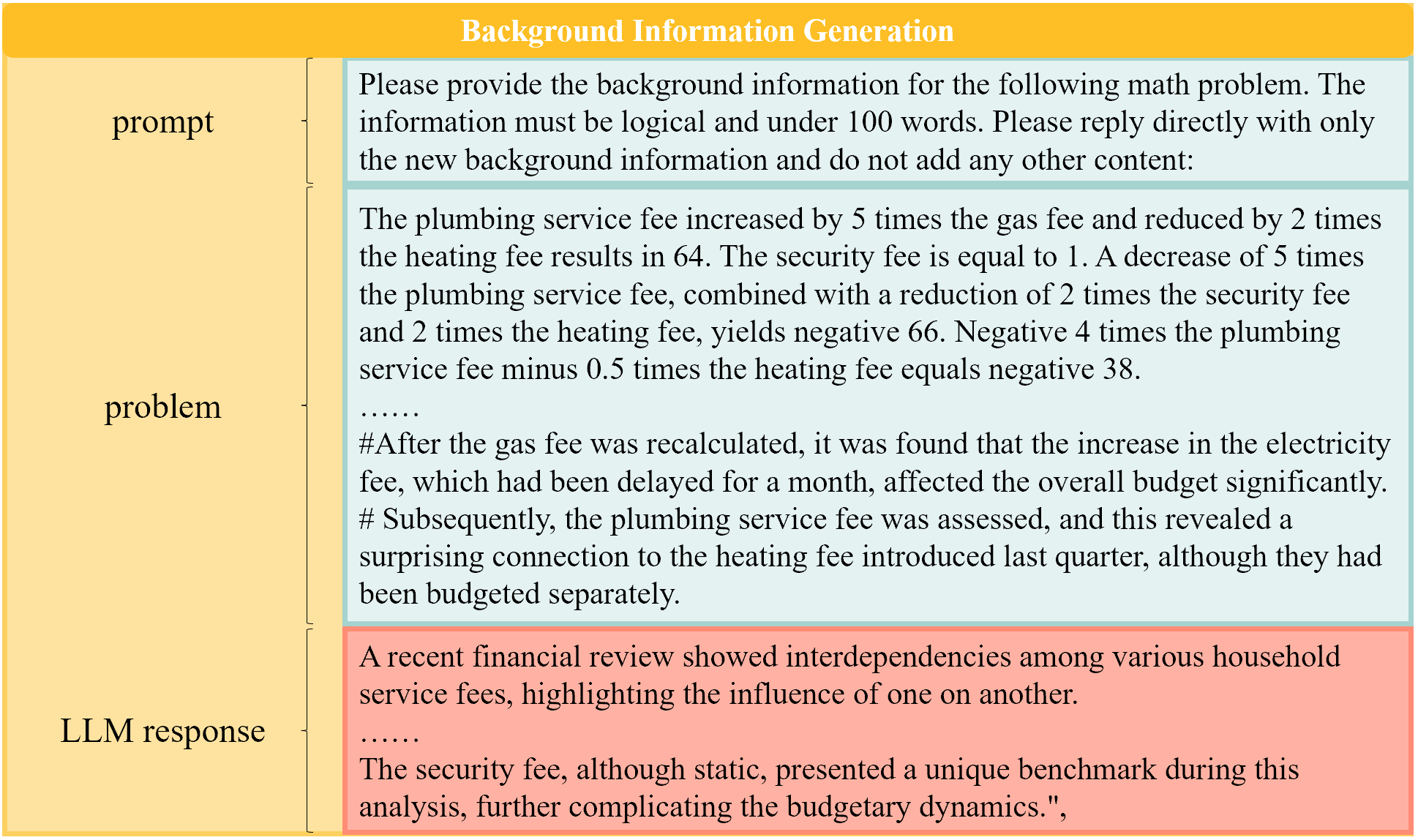}
    \caption{Background Information Generation}
    \label{fig:Background_Information_Generation}
\end{figure*}

\begin{figure*}[htbp]
    \centering
    \includegraphics[width=\textwidth]{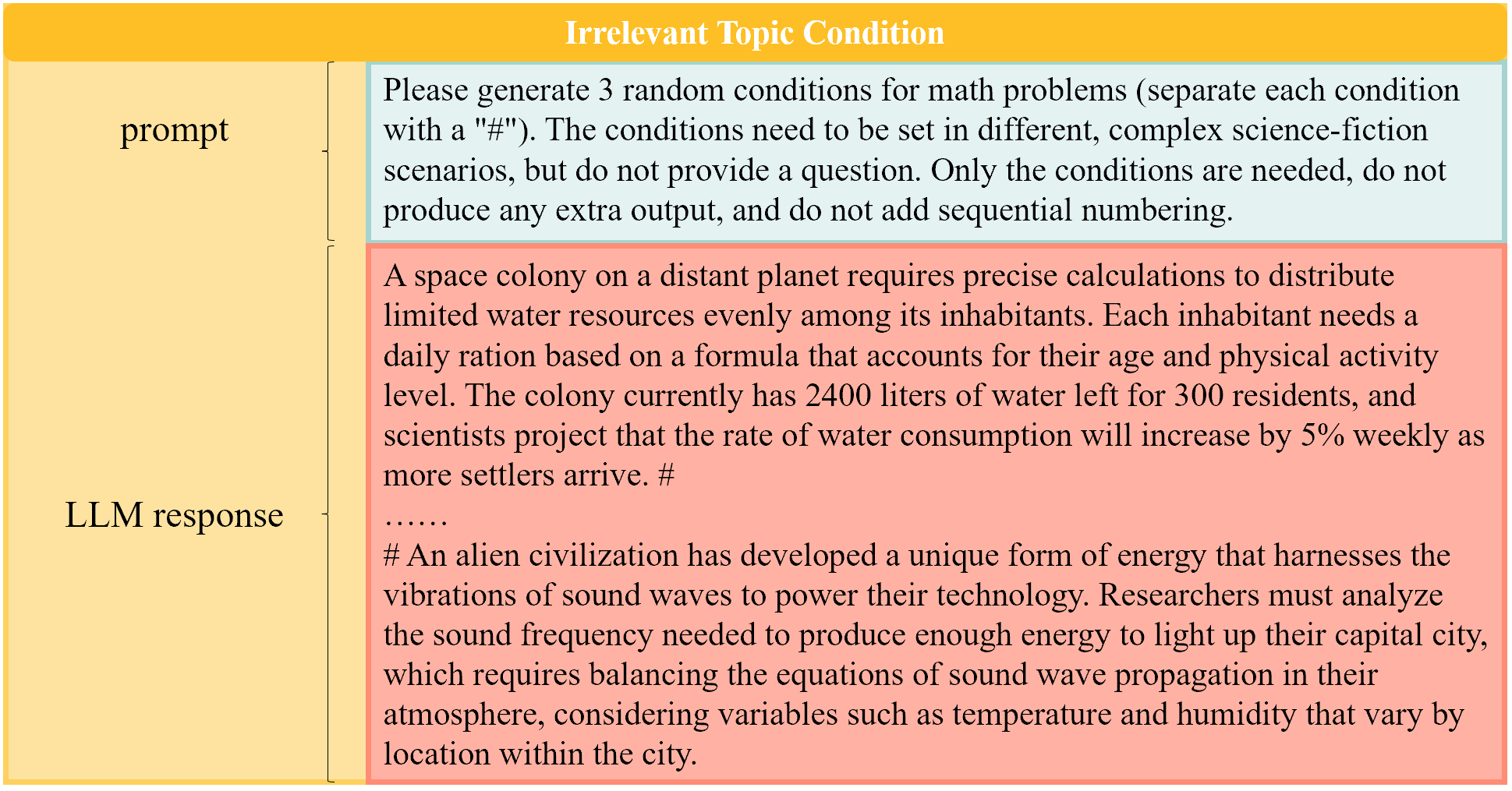}
    \caption{Irrelevant Topic Condition}
    \label{fig:Irrelevant_Topic_Condition}
\end{figure*}

Introducing misleading information at the linguistic level is a critical component of the workflow, achieved through the following:
\begin{itemize}
    \item \textbf{Misleading Textual Condition}: 
    Adds linguistically deceptive descriptions (see Figure~\ref{fig:Misleading_Textual_Condition}).
    \item \textbf{Background Information Generation}: 
    Creates relevant background contexts for the problems (see Figure~\ref{fig:Background_Information_Generation}).
    \item \textbf{Irrelevant Topic Condition}: 
    Introduces distracting conditions unrelated to the problem's solution (see Figure~\ref{fig:Irrelevant_Topic_Condition}).
\end{itemize}

The \textbf{Evolution and Evaluation Stage} concludes the process:
\begin{itemize}
    \item \textbf{Crossover Operator}: 
    Crosses pairs of generated problems to produce new problem variants (as described in Algorithm~\ref{alg:Crossover_Operator}).
    \item \textbf{Fitness Function}: 
    Calculates a fitness score to evaluate the quality of each problem, thereby determining the problems that will proceed to the next evolutionary round (see Algorithm~\ref{alg:Fitness_Function}).
\end{itemize}

\begin{algorithm*}[htbp]
\caption{Crossover Operator}
\label{alg:Crossover_Operator}
\begin{algorithmic}[1]
    \State $NewDataSet \gets [~]$

    \State Create pairs of problems (A, B) from the initial `DataSet`.
    
    \For{each created pair (Problem A, Problem B)}
        \State Create two new chained problems from each pair: A→B and B→A.
        \State For each direction (e.g., from A to B):
        \State \quad Calculate a `relation\_ratio` connecting the answer of A to a target number in B's text.
        \State \quad Create a `linking\_instruction` text that describes this ratio.
        \State \quad Modify B's text by replacing the target number with a placeholder 'X'.
        \State \quad Assemble a `new\_prompt` by combining A's text, the instruction, and the modified B's text.
        \EndFor
\end{algorithmic}
\end{algorithm*}

\begin{algorithm*}[htbp]
\caption{Fitness Function}
\label{alg:Fitness_Function}
\begin{algorithmic}[1]
    \State $DataSet \gets$ Read problem data from $inputFile$.
    \State Pre-process all problem prompts using a natural language processing model (e.g., spaCy).

    \State Initialize an empty list 'AllFeatures`.
    \For{each $Problem$ in $DataSet$}
        \State Calculate a feature set for the $Problem$, including:
        \Statex \quad \quad • \textbf{Linguistic Features:} \texttt{word\_count}, \texttt{readability\_score}, \texttt{syntactic\_complexity}, \texttt{word\_entropy}.
        \Statex \quad \quad • \textbf{Mathematical Features:} \texttt{num\_equations}, \texttt{num\_variables}, \texttt{non\_linear\_count}, \texttt{noise\_ratio}.
        \Statex \quad \quad • \textbf{Dataset-level Features:} \texttt{second\_highest\_similarity\_score}.
        \Statex \quad \quad • \textbf{Reference Score:} \texttt{referee\_difficulty\_score}.
        \State Append the calculated feature set to `AllFeatures`.
    \EndFor

    \State Assemble all feature sets from 'AllFeatures` into a numerical matrix, $M$.
    \State Apply Min-Max normalization to matrix $M$ to get a normalized matrix, $M_{norm}$.
    \State Define a predefined `weights` vector for all features.
    \State Calculate the final \texttt{difficulty\_scores} by taking the dot product of $M_{norm}$ and the `weights` vector, then scale the result (e.g., to a 0-10 range).
\end{algorithmic}
\end{algorithm*}

% \textbf{Genetic Operators: }

% \begin{algorithm*}[htbp]
% \caption{Approximate Replacement}
% \label{alg:Approximate_Replacement}
% \begin{algorithmic}[1]
%     \State $DataSet \gets$ Read JSON data from $inputFile$.
%     \State Define $ConfusionMap$ of ambiguous symbols.

%     \For{each $Item$ in $DataSet$}
%         \State Let $InitialEquations$ be the list of equations in the item's prompt.
%         \If{a directly-assignable variable ($v$) can be found in $InitialEquations$ AND the ``Templated Reconstruction" strategy using $v$ succeeds}
%             \State $confused\_eq \gets$ The result of the successful reconstruction.
%         \Else
%             \State $confused\_eq \gets$ The result of applying ``Random Interference" to the last equation.
%         \EndIf
        
%         \State Append the generated $confused\_eq$ to the item's prompt.
%     \EndFor
    
%     \State Save the modified $DataSet$ to $outputFile$.
% \end{algorithmic}
% \end{algorithm*}

% \begin{figure*}[htbp]
%     \centering
%     \includegraphics[width=\textwidth]{figures/Useless_Mathematical_Condition.png}
%     \caption{Useless Mathematical Condition}
%     \label{fig:Useless_Mathematical_Condition}
% \end{figure*}

% \begin{figure*}[htbp]
%     \centering
%     \includegraphics[width=\textwidth]{figures/Misleading_Mathematical_Condition.png}
%     \caption{Misleading Mathematical Condition}
%     \label{fig:Misleading_Mathematical_Condition}
% \end{figure*}

% --- Appendix A: Process Overview (Revised) ---
\section{Detailed Methodology for Seed Problem Initialization}
\label{app:Seed_Problem}
This appendix provides a detailed breakdown of the methodology used in the Seed Problem Initialization stage (Section~\ref{Seed_Problem_Initialization}). 
The primary goal of this stage is to generate an initial set of mathematical problems. 
These problems must be logically rigorous and guaranteed to have a unique integer solution.
Additionally, their structure must emulate the step-by-step reasoning required for authentic word problems.
To achieve this, our framework employs a reverse-engineering strategy.

The process is comprised of three main steps:

\subsection{Benchmark Solution Pre-setting}
The generation process begins with the answer. 
A benchmark solution vector, $\vec{s}$, is randomly generated within a predefined integer space (e.g., integers in $[1, 20]$). 
By pre-setting the solution, we fundamentally guarantee that every generated seed problem has a determinate and unique integer answer.

\subsection{Sparse Equation Construction}
To mimic real-world word problems, we construct a sparse linear system of equations ($A\vec{x} = \vec{b}$). 
"Sparse" simply means that each equation uses only a few of the total variables (e.g., two per equation).

This design encourages the model to perform \textbf{step-by-step reasoning}, much like a person would, rather than solving a single, complex matrix all at once. 
After creating the sparse matrix $A$ and the solution vector $\vec{s}$, we find the constants for the problem by calculating $\vec{b} = A\vec{s}$.

\subsection{Algebraic Quality Assurance}
To guarantee the logical integrity and quality of each generated problem, every system of equations undergoes two critical automated checks before being accepted into the initial population:

\textbf{Full Rank Check:} This check verifies that the coefficient matrix $A$ is of full rank (i.e., $\text{rank}(A) = n$ for an $n \times n$ system). From linear algebra, this condition guarantees that the system of equations has a single, unique solution. This step is crucial for eliminating ambiguity and ensuring that each problem is well-posed.
    
\textbf{Necessity Check:} This check confirms that there are no redundant equations. 
This is particularly critical for the effectiveness of the Crossover Operator (Section~\ref{Crossover_Operator}), as it guarantees that its modifications will meaningfully impact the problem's solution path. 
The detailed mathematical formulation for this check is provided in Appendix~\ref{app:necessity_proof}.

Through this multi-stage, validation-driven process, EvolMathEval produces a high-quality and reliable initial population of seed problems. 
Each problem is certified to be logically sound and uniquely solvable, providing a robust foundation for the subsequent evolutionary operations.

% --- Appendix B: Mathematical Proof (New) ---
\section{Detailed Design of the Necessity Check}
\label{app:necessity_proof}

This appendix details the mathematical basis for the Necessity Check, a validation step that guarantees the integrity of our Crossover Operator. 
The operator creates a long-chain reasoning task by replacing the first value in the first equation of a second problem with a placeholder 'X'.
This value is derived from the first problem's solution.

The check prevents a critical failure. 
If the modified first equation were mathematically redundant, a solver could simply ignore it and bypass the first problem entirely. 
This action would render the entire crossover operation useless.
The Necessity Check confirms this equation is indispensable, thereby forcing the intended logical dependency. 
This check is performed after a Full Rank Check has already guaranteed the system has a unique solution.

\subsection{Defining Prerequisite Variables}
Consider a system of $k$ equations where the target variable is $x_k$. 
The solution for $x_k$ is derived from the $k$-th equation: $A_{k,1}x_1 + \dots + A_{k,k}x_k = b_k$. 
To solve for $x_k$, the values of all other variables with non-zero coefficients in this equation must be determined first. 
We define this set of variables as the \textbf{Prerequisite Variables}, denoted $V_{\text{preq}}$, with their corresponding column indices being the set $J$.
% $$ V_{\text{preq}} = \{x_j \mid A_{k,j} \neq 0, \quad 1 \le j < k\} $$
% $$ J = \{j \mid x_j \in V_{\text{preq}}\} $$
\begin{equation}
    V_{\text{preq}} = \{x_j \mid A_{k,j} \neq 0, \quad 1 \le j < k\}
\end{equation}
\begin{equation}
    J = \{j \mid x_j \in V_{\text{preq}}\}
\end{equation}
The cardinality of this set is denoted by $d = |V_{\text{preq}}|$.

\subsection{Constructing the Validation Sub-matrix}
To test if the first equation is necessary for solving the variables in $V_{\text{preq}}$, we construct a \textbf{Validation Sub-matrix}, $A_{\text{val}}$, using the equations excluding the first one (i.e., equations $2, \dots, k-1$). 
This sub-matrix is formed by selecting the rows $2, \dots, k-1$ and the columns corresponding to the indices in $J$ from the original coefficient matrix $A$.
The resulting matrix $A_{\text{val}}$ has dimensions $(k-2) \times d$, formally defined as:
% \[
% A_{\text{val}} = (A_{i,j})
% \]
\begin{equation}
    A_{\text{val}} = (A_{i,j})
\end{equation}
\noindent where $i \in \{2, \dots, k-1\}$ and $j \in J$.

\subsection{Applying the Decision Criterion}
The decision to accept or reject the problem is based on the rank of the validation sub-matrix $A_{\text{val}}$ relative to the number of prerequisite variables, $d$.

\begin{itemize}
    \item \textbf{Redundancy Condition (Reject):} 
    If $\text{rank}(A_{\text{val}}) = d$, it implies that equations $2$ through $k-1$ are sufficient on their own to uniquely determine all prerequisite variables. 
    In this scenario, the first equation is redundant for solving the target variable. 
    The problem instance is therefore \textbf{rejected}.

    \item \textbf{Necessity Condition (Accept):} 
    If $\text{rank}(A_{\text{val}}) < d$, it implies that equations $2$ through $k-1$ are insufficient to uniquely determine all prerequisite variables. 
    The system relies on the additional constraints from the first equation to find a solution. 
    Since the entire system has already passed the Full Rank Check, this condition confirms the \textbf{necessity} of the first equation. 
    The problem instance is therefore \textbf{accepted}.
\end{itemize}

\section{End-to-End Example of the Evolutionary Process}
\label{app:full_example}

\begin{figure*}[h!]
    \centering
    % 重要：请将下面的 "figures/full_evolution_example.png" 替换为你实际的图片路径和文件名
    \includegraphics[width=\textwidth, trim=0cm 15.1cm 0cm 0cm, clip]{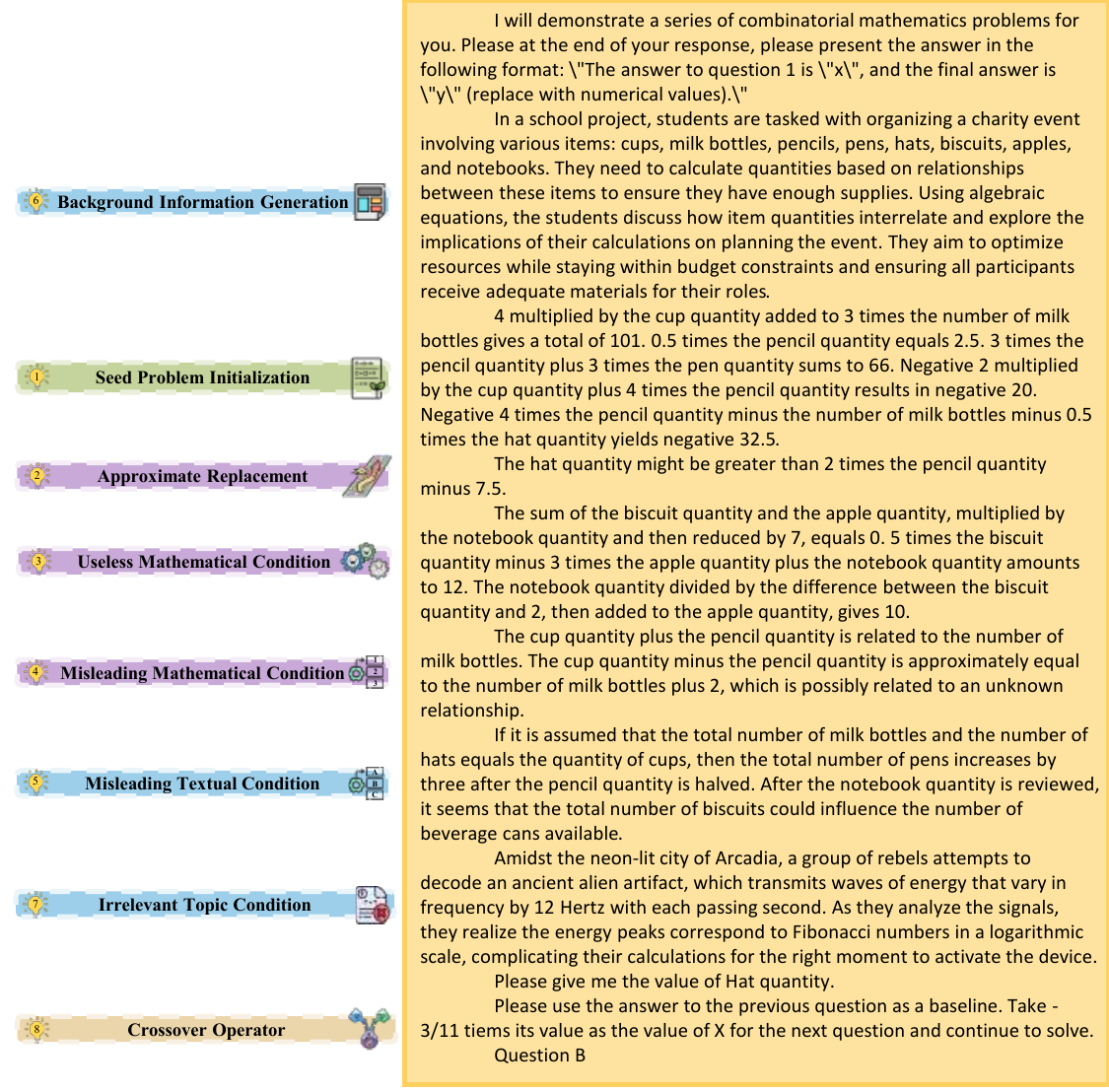}
    \caption{A showcase of outputs from the core evolutionary operators. The text block on the right serves as an independent example for the corresponding operator named on the left.}
    \label{fig:full_evolution_example}
\end{figure*}

To provide concrete illustrations of our framework's core operators, Figure~\ref{fig:full_evolution_example} presents a showcase of their individual outputs. 
Each numbered section pairs an operator's name on the left with an independent example of the text it generates on the right. 
This serves to visually anchor the abstract descriptions of each operator provided in the main methodology.

\section{Configuration Parameters}
\label{appendix:I}

% Our experimental parameters are listed in Table~\ref{tab:config_params}. 
% The \textbf{Selection Fitness Threshold} controls the final difficulty: problems scoring below this value are re-evolved, so a higher threshold results in a more challenging benchmark.

Our experimental parameters are listed in Table~\ref{tab:config_params}. 
The \textbf{Selection Fitness Threshold} is a key adjustable hyperparameter that controls the final difficulty. 
Problems scoring below this value are re-evolved, so a higher threshold can be set to produce a more challenging benchmark.

\begin{table}[h!]
\centering
\resizebox{\columnwidth}{!}{%
\begin{tabular}{@{}cc@{}} % 第一个 l 代表第一列左对齐，c 代表第二列居中
\toprule
\textbf{Parameter} & \textbf{Value} \\ \midrule
\quad Population Size & 300 (EvolMath-300) \\
\quad Extended Population Size & 1,000 (EvolMath-1000) \\
\quad Selection Fitness Threshold & -0.5 \\
\quad Number of Generations & 2 \\
\quad Variables per Problem & 5 \\
\quad Equations per Problem & 5 \\
\quad Solution Space & Integers in [1, 20] \\
\quad Equation Sparsity & 2 variables per equation \\ \bottomrule
\end{tabular}
}
\caption{Key Configuration Parameters for the EvolMathEval Framework.}
\label{tab:config_params}
\end{table}

\section{The Stability of LLMs' Performance on EvolMath}
\label{appendix:F}
To assess the performance stability of LLMs on the EvolMath benchmark, we selected three models and conducted four repeated experiments for each. 
The results confirmed that their performance is stable. See Table~\ref{tab:accuracy_stability}.

The output of Large Language Models can have a degree of randomness. 
So it is essential to confirm the stability of their performance. 
To ensure our findings on EvolMath are reliable and not the result of chance, we conducted a stability analysis. 
We selected three models representing different performance tiers and ran four repeated evaluations for each.

The results, presented in Table~\ref{tab:accuracy_stability}, demonstrate remarkable consistency.
For example, the four final accuracy scores for DeepSeek-R1 (79.03, 81.33, 80.49, 79.89) show a standard deviation of just 1.01, indicating very low variance. 
A similar stability is observed for the other models. 
This confirms that EvolMath elicits a stable and reproducible measure of a model's capabilities, thereby strengthening the validity of our primary conclusions.

\begin{table}[h!]
\centering
\begin{tabular}{l c c}
\toprule
\textbf{Model} & \textbf{Q1} & \textbf{Final}\\
\midrule
\multirow{4}{*}{DeepSeek-R1}    & 87.10  & 79.03  \\
                                & 88.56  & 81.33  \\
                                & 87.74 & 80.49 \\
                                & 89.30 & 79.89 \\
\midrule % Separator line
\multirow{4}{*}{Doubao-Seed-1.6} & 83.33  & 72.70  \\
                                & 79.79  & 72.32  \\
                                & 82.02 & 74.49 \\
                                & 81.05 & 70.53 \\
\midrule % Separator line
\multirow{4}{*}{Qwen3-235b-a22b} & 31.19  & 18.35 \\
                                & 32.86  & 20.00  \\
                                & 34.76  & 20.00 \\
                                & 34.76  & 21.53 \\
\bottomrule
\end{tabular}
\caption{Stability of Model Accuracy (\%) on EvolMath Across Four Repeated Experiments.}
\label{tab:accuracy_stability}
\end{table}

\section{Other Prompts}
\label{appendix:C}
\begin{figure*}[htbp]
    \centering
    \includegraphics[width=\textwidth]{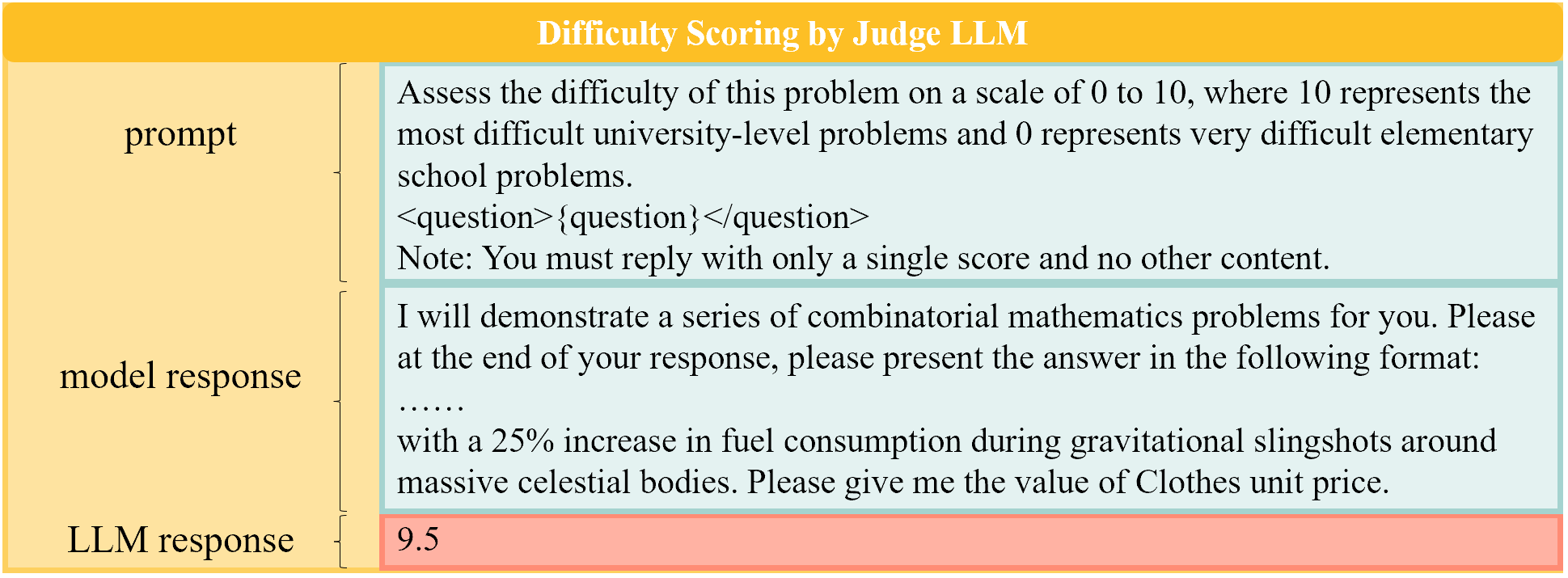}
    \caption{Difficulty Scoring by Judge LLM}
    \label{fig:Difficulty_Scoring_by_Judge_LLM}
\end{figure*}

\begin{figure*}[htbp]
    \centering
    \includegraphics[width=\textwidth]{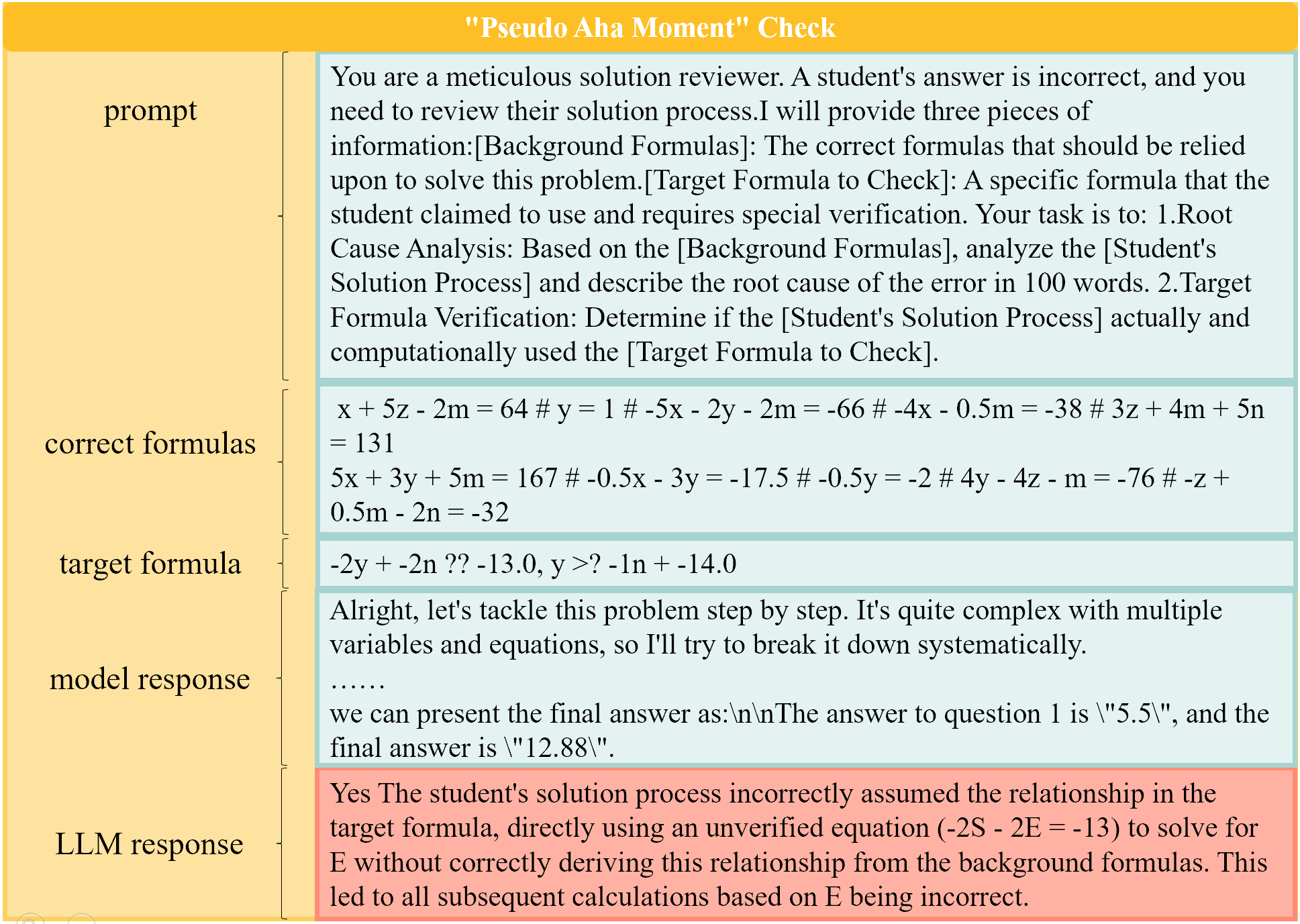}
    \caption{Pseudo Aha Moment Check}
    \label{fig:Pseudo_Aha_Moment_Check}
\end{figure*}

This section presents the additional prompts we utilized. Specifically, we detail the prompt employed within the Fitness Function to score problems (see Figure~\ref{fig:Difficulty_Scoring_by_Judge_LLM}) and the prompts used to examine the proportion of Pseudo Aha Moment and analyze the causes of incorrect solutions (see Figure~\ref{fig:Pseudo_Aha_Moment_Check}).

\section{Fitness Function}
\label{appendix:H}
% This principle is mathematically captured in the following equation for the comprehensive fitness score ($S$):
% \begin{equation}
% S = - \sum_{i=1}^{M} \frac{r_i (1 - p_i)}{\sum_{j=1}^{M} |r_j (1 - p_j)|} \cdot f_i
% \label{eq:comprehensive_fitness}
% \end{equation}
% In this formula, for each feature $i$, $f_i$ is its value, $r_i$ is its Pearson correlation coefficient with model accuracy, and $p_i$ is its t-test p-value.

This appendix provides the specific mathematical formula for calculating the data-driven weight ($w_i$) for each feature in fitness function, as introduced in Section~\ref{sec:validation_fitness}.
This method is designed to ensure that a feature's weight accurately reflects its measured impact on problem difficulty.

The weight for each feature $i$ is defined as:
\begin{equation}
w_i = \frac{r_i (1 - p_i)}{\sum_{j=1}^{M} |r_j (1 - p_j)|}
\label{eq:weight_calculation}
\end{equation}
% This formula directly implements our data-driven principles. The numerator, $r_i (1 - p_i)$, ensures that a feature's weight magnitude is large only when it has both a strong correlation with model accuracy (high $|r_i|$) and high statistical significance (low p-value, $p_i$). The denominator is a normalization constant, ensuring that the absolute values of all weights sum to 1. The sign of the weight is determined by the correlation direction ($r_i$).
% In this formula, for each feature $i$, $r_i$ is its Pearson correlation coefficient with model accuracy, and $p_i$ is its t-test p-value.

These weights are then used to compute the comprehensive fitness score ($S$) as a weighted sum of the normalized feature values ($f_i$):
\begin{equation}
S = - \sum_{i=1}^{M} w_i \cdot f_i
\label{eq:comprehensive_fitness}
\end{equation}
% The negative sign aligns the final score with our definition of difficulty. For instance, a feature that makes a problem easier will have a positive correlation ($r_i > 0$) and thus a positive weight ($w_i$). This positive weight, when multiplied by the negative sign, correctly decreases the overall difficulty score $S$.

In these formulas, for each feature $i$:
$f_i$ is its normalized value.
$r_i$ is its Pearson correlation coefficient with model accuracy.
$p_i$ is its t-test p-value.

% \section*{D. Details of Large Language Models}
% \label{appendix:D}

% This section provides details for the LLMs used in our experiments.

% \begin{table}[hbt!]
% \centering
% \caption{Details of the Large Language Models tested. Models marked with an asterisk (*) are assumed to be hypothetical or future versions based on their names.}
% \label{tab:llm_details}
% \begin{tabular}{lll}
% \toprule
% \textbf{Model} & \textbf{Parameters} & \textbf{Notes} \\
% \midrule
% DeepSeek-V3 & Not Specified & \\
% DeepSeek-R1 & Not Specified & \\
% Doubao-Seed-1.6 & Not Specified & Model from ByteDance. \\
% Gemini-2.5-Pro* & Not Specified & Assumed to be a hypothetical future model. \\
% GLM-4-Flash & Not Specified & A lightweight, fast version of the GLM-4 model. \\
% GPT-o3 & Not Specified & Assumed to be a typo for GPT-4o. \\
% Kimi-K2 & Not Specified & Model from Moonshot AI, known for long context. \\
% Llama-3-70b & 70 Billion & Model from Meta. \\
% Llama-3-8b & 8 Billion & Model from Meta. \\
% Qwen3-235b-a22b* & 235B (22B active) & Assumed to be a hypothetical MoE model. \\
% Qwen3-30b-a3b* & 30B (3B active) & Assumed to be a hypothetical MoE model. \\
% Qwen3-32b* & 32 Billion & Assumed to be a hypothetical dense model. \\
% \bottomrule
% \end{tabular}
% \end{table}

\section{Comparison of EvolMath with Other Benchmarks}
\label{appendix:G}

% As shown in Table~\ref{tab:appendix:G}, a comparison of model performance on several benchmarks reveals that EvolMath is significantly more challenging than both AMWPG and GSM-PLUS.

To measure the difficulty of EvolMath, we compare its performance against two other challenging math reasoning datasets: AMWPG and GSM-PLUS.

The results, detailed in Table~\ref{tab:appendix:G}, show that model accuracies are substantially lower on EvolMath. 
This highlights its significant challenge to even the most capable large language models.

\begin{table}[t!] % [t!] 建议 LaTeX 将其放在页面顶部
\centering
\resizebox{\columnwidth}{!}{%
\begin{tabular}{l cc cc cc}
\toprule
\textbf{Model} & \textbf{EvolMath} & \textbf{AMWPG} & \textbf{GSM-PLUS} \\
\midrule
DeepSeek-R1 & 79.03 & 91.67 & 96.00 \\
DeepSeek-V3 & 21.51 & 94.00 & 90.00 \\
Doubao-Seed-1.6 & 72.70 & 92.00 & 96.00 \\
Gemini-2.5-Pro & 76.47 & 88.00 & 94.00 \\
GLM-4-Flash & 2.89 & 62.00 & 64.00 \\
Kimi-K2 & 40.71 & 94.00 & 92.00 \\
GPT-o3 & 65.55 & 94.00 & 96.00 \\
Qwen3-30b-a3b & 13.09 & 83.33 & 94.00 \\
Qwen3-32b & 6.73 & 88.00 & 92.00 \\
Qwen3-235b-a22b & 18.35 & 86.00 & 89.80 \\
\bottomrule
\end{tabular}
}
\caption{Model Accuracy (\%) on EvolMath, AMWPG, and GSM-PLUS.}
\label{tab:appendix:G}
\end{table}

\section{Performance on Evolved Public Benchmarks}
\label{appendix:D}
% We quantitatively evaluated the effectiveness of EvolMathEval on multiple public datasets.
% The main body of the paper primarily shows the percentage drop in accuracy for each model on three datasets. 
% Detailed data is presented in Table~\ref{tab:wide_table_example}.

This appendix provides the full data for our generalization experiments, corresponding to the analysis in Section~\ref{Validation_of_Generalization_Capability}.

While the main paper's Table~\ref{tab:accuracy_drop} presents a summary metric (the relative percentage drop in accuracy), this section provides the foundational data for that calculation. Table~\ref{tab:wide_table_example} details the complete absolute accuracy scores (\%) for each model on the original and evolved public benchmarks.

\begin{table*}[t!] % [t!] 建议 LaTeX 将其放在页面顶部
\centering
\begin{tabular}{l cc cc cc}
\toprule
\textbf{Model} & \multicolumn{2}{c}{\textbf{GSM8K}} & \multicolumn{2}{c}{\textbf{SVAMP}} & \multicolumn{2}{c}{\textbf{MAWPS}} \\
\cmidrule(lr){2-3} \cmidrule(lr){4-5} \cmidrule(lr){6-7}
 & \textbf{Original} & \textbf{Evolved} & \textbf{Original} & \textbf{Evolved} & \textbf{Original} & \textbf{Evolved} \\
\midrule
Doubao-Seed-1.6 & 96.50 & 55.15 & 97.50 & 71.73 & 95.50 & 64.02 \\
Kimi-K2 & 95.95 & 54.17 & 98.18 & 63.27 & 93.22 & 62.00 \\
DeepSeek-R1 & 98.65 & 51.72 & 96.00 & 71.79 & 92.00 & 59.38 \\
DeepSeek-V3 & 97.00 & 37.16 & 96.00 & 52.22 & 94.00 & 42.78 \\
Gemini-2.5-Pro & 96.00 & 62.00 & 97.33 & 71.43 & 92.00 & 64.00 \\
GLM-4-Flash & 85.00 & 4.52 & 90.50 & 17.69 & 92.00 & 8.38 \\
GPT-o3 & 97.00 & 60.00 & 96.00 & 73.47 & 93.33 & 62.00 \\
Llama-3-70b & 90.24 & 29.08 & 87.50 & 42.42 & 92.68 & 36.59 \\
Llama-3-8b & 78.57 & 10.71 & 83.87 & 28.57 & 91.89 & 12.50 \\
Qwen3-235b-a22b & 96.00 & 50.00 & 100.00 & 50.00 & 93.33 & 56.00 \\
Qwen3-30b-a3b & 96.50 & 43.94 & 97.50 & 65.83 & 94.90 & 52.02 \\
Qwen3-32b & 97.50 & 44.67 & 98.00 & 63.96 & 95.79 & 55.56 \\
\bottomrule
\end{tabular}
\caption{Performance on Evolved Public Benchmarks.}
\label{tab:wide_table_example}
\end{table*}

\section{Detailed Analysis of a "Pseudo Aha Moment" Failure Case}
\label{appendix:E}

\begin{figure*}[htbp]
    
    \includegraphics[width=\textwidth, trim=0cm 5.9cm 0cm 0cm, clip]{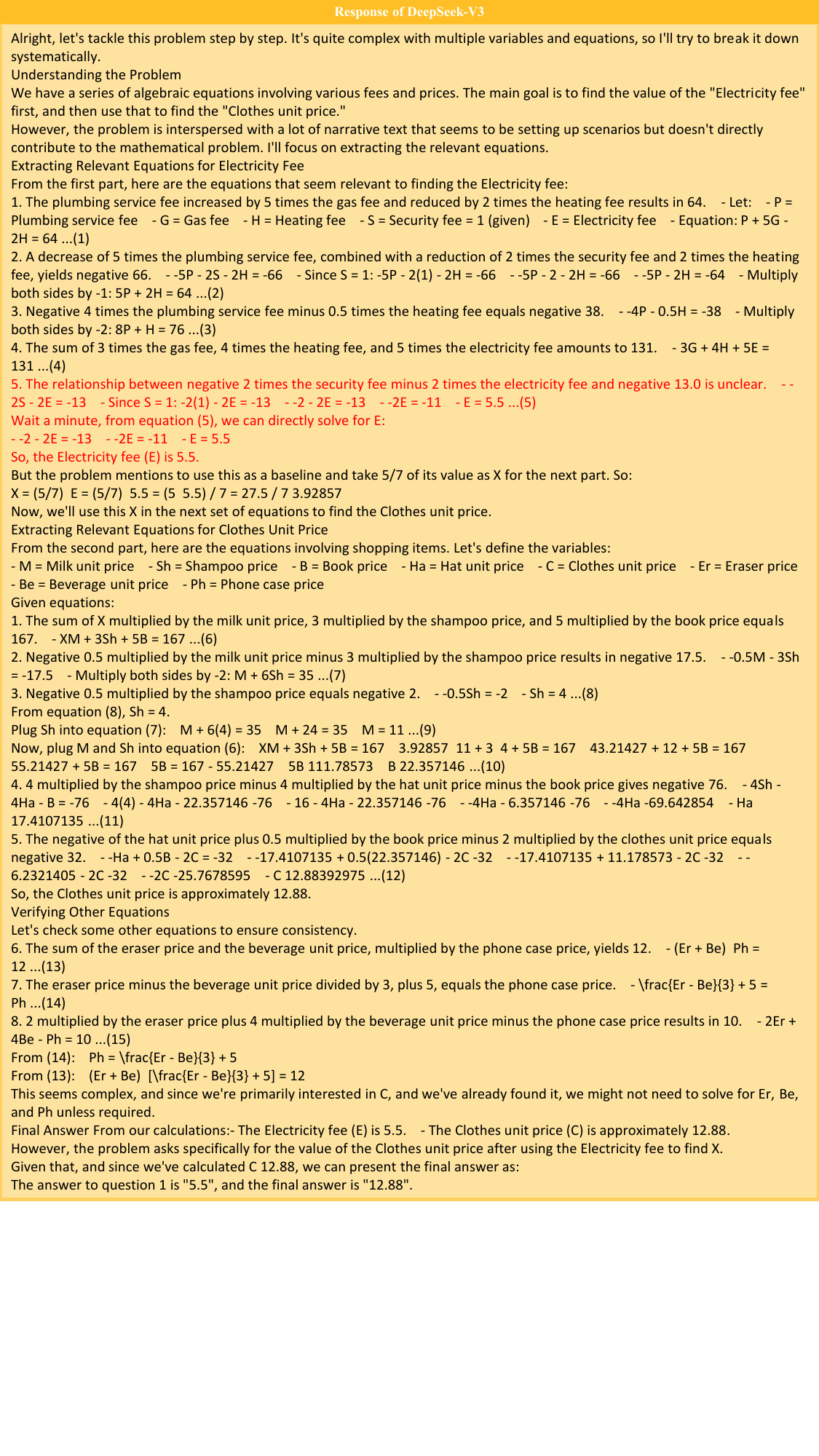}
    \caption{Reasoning trace from DeepSeek-V3 showing the "Pseudo Aha Moment" phenomenon. The response follows a flawed shortcut, highlighted in red.}
    \label{fig:response}
\end{figure*}

This appendix dissects the provided response (Figure~\ref{fig:response}) to illustrate the ``Pseudo Aha Moment'' failure. 
The analysis is broken down into three key stages:

\textbf{The Flawed Shortcut:} 
    The critical error, highlighted in red in the figure, occurs at the fifth condition in the initial problem. 
    The model notes, \texttt{``Wait a minute... we can directly solve for E,''} by seizing upon a reasoning shortcut. 
    Its crucial mistake is ignoring that the condition was an \textbf{approximation} ($-2S - 2E \approx -13$) and treating it as a \textbf{precise, solvable equation}. 
    This confident leap to an easy but invalid shortcut is the ``Pseudo Aha Moment.''

\textbf{Bypassing Rigorous Reasoning:} 
    By taking this shortcut, the model bypasses the required multi-step reasoning path of solving the first four equations of the initial problem. 
    This reveals a strong bias for computational simplicity over logical fidelity, as it avoids the complex but correct path in favor of an easy but flawed one.

\textbf{Error Propagation:} 
    The initial error from misinterpreting the approximation then propagates. 
    The incorrect value of $E$ leads to a faulty linking variable $X$. 
    Consequently, even with flawless subsequent algebra, the final answer is guaranteed to be wrong because it is built upon this flawed premise.

% This is a complete example of a model's incorrect response resulting from the flawed application of a vague reasoning shortcut, see Figure~\ref{fig:response}.

% \section{Limitations and Future Work}
% \label{sec:limitations_future_work}

% \subsection{Limitations}
% Despite its demonstrated effectiveness, our framework has two primary limitations:

% \begin{itemize}
% \item \textbf{Domain Specificity:} The framework's current implementation focuses on generating algebra-based mathematical reasoning problems. This limits its direct applicability to other important mathematical domains that require different logical representations, such as geometry, probability, or calculus.

% \item \textbf{Indirectness of Difficulty Assessment:} The fitness function, while effective, serves as a proxy for true problem difficulty. It quantitatively captures structural and linguistic complexity but may not fully model the implicit cognitive challenges posed by deep semantic ambiguities or novel logical traps that would challenge a human reasoner.
% \end{itemize}

\section{Limitations and Future Work}
\label{sec:limitations_future_work}

\subsection{Limitations}
Despite its demonstrated effectiveness, our framework has two primary limitations:

\begin{itemize}
    \item \textbf{Domain Specificity:} The framework's current implementation focuses on generating algebra-based mathematical reasoning problems. 
    This limits its direct applicability to other important mathematical domains that require different logical representations, such as geometry, probability, or calculus.

    \item \textbf{Indirectness of Difficulty Assessment:} 
    The fitness function, while effective, serves as a proxy for true problem difficulty. 
    It quantitatively captures structural and linguistic complexity.
    However, it may not fully model the implicit cognitive challenges posed by deep semantic ambiguities or novel logical traps that would challenge a human reasoner.
\end{itemize}

\subsection{Future Work}
Future work will focus on two main directions to build upon the foundation of EvolMathEval:

\begin{itemize}
    \item \textbf{Generalization to New Domains:} 
    A key priority is to generalize the framework's evolutionary mechanisms beyond algebra. 
    We plan to adapt the seed generation and genetic operators for other complex reasoning domains, such as code generation, logical puzzle solving, and geometric reasoning. 
    This will allow us to create a more comprehensive suite of evolvable evaluation benchmarks.

    \item \textbf{Enhancing Model Robustness:} 
    The high-difficulty problems generated by EvolMathEval form an ideal dataset for improving LLM mathematical reasoning.
    We intend to use these evolved samples to fine-tune existing models. 
    This targeted training aims to directly address the identified weaknesses, such as the "Pseudo Aha Moment." 
    By doing so, it enhances the models' logical robustness and reduces their reliance on flawed heuristics.
\end{itemize}

\end{document}